\documentclass[11pt,letterpaper]{article}
\usepackage[T1]{fontenc}
\usepackage[utf8]{inputenc}
\usepackage[margin=1in]{geometry}
\usepackage{amsmath,amssymb}
\usepackage{newtxtext,newtxmath}
\usepackage{bm}
\usepackage{booktabs,tabularx,threeparttable,array,makecell}
\usepackage{graphicx}
\usepackage{microtype}
\usepackage{enumitem}
\usepackage{placeins}
\usepackage{longtable}
\usepackage{pdflscape}
\usepackage{caption}
\usepackage[round,authoryear]{natbib}
\usepackage{xurl}
\usepackage[hidelinks]{hyperref}

\setlist{nosep,leftmargin=*}
\newcolumntype{Y}{>{\raggedright\arraybackslash}X}
\newcolumntype{P}[1]{>{\raggedright\arraybackslash}p{#1}}

\newcommand{\NA}{\textemdash}
\title{When Can LLM Digital Twins Reduce Human Measurement? From Behavioral Fidelity to Statistical Substitutability}
\hypersetup{pdftitle={When Can LLM Digital Twins Reduce Human Measurement? From Behavioral Fidelity to Statistical Substitutability}}
\author{%
  Steven Wang \quad Kyle Hunt \quad Shaojie Tang \quad Kenneth Joseph\\[0.4em]
  \normalsize University at Buffalo
}
\hypersetup{pdfauthor={Steven Wang, Kyle Hunt, Shaojie Tang, Kenneth Joseph}}

\date{}

\begin{document}
\maketitle
\begin{abstract}
LLM-based digital twins promise to reduce repeated human data collection by generating person-specific responses, yet existing evaluations provide little evidence about whether they can reduce human measurement while preserving valid inference. To address this, we introduce \textit{statistical substitutability}, an inferential criterion that evaluates the extent to which twin predictions can reduce human measurement for a particular estimand while preserving valid inference. We develop a framework, grounded in mixed-subject and prediction-powered inference, that evaluates statistical substitutability along four dimensions: aggregate fidelity, paired respondent-level signal, finite-sample human-label recovery, and stability across populations. Across two empirical evaluations spanning behavioral experiments, multiple models, and alternative respondent representations, we find that digital twins can reproduce average human effects while providing little information about which individuals differ from those averages. Newer models and richer respondent information improve some dimensions of performance but do not reliably translate into human-data savings. Human calibration can reduce aggregate prediction error, yet limited labeled samples often fail to produce stable precision gains. Importantly, these findings demonstrate that behavioral fidelity is neither necessary nor sufficient for statistical substitutability. More broadly, they suggest that AI-generated evidence should be evaluated based on its ability to support valid scientific inference rather than its ability to reproduce human outcomes alone. Digital twins should therefore be judged for confirmatory use by whether they reduce uncertainty about human quantities, not merely by whether they reproduce human means, distributions, or effects.
\end{abstract}

\section{Introduction}

Generative AI now supports literature synthesis \citep{asai2026}, coding \citep{merow2023}, measurement \citep{halterman2026}, simulation \citep{argyle2023}, experimental design \citep{boiko2023}, and analysis \citep{ziems2024}. These tools can expand the scale and speed of research, but they also raise questions about validity, provenance, verification, and human responsibility. \citet{gopal2025} describe this shift as a change in scholarly infrastructure and argue that AI should serve as a research partner rather than a research driver. We study one concrete use of this infrastructure, namely whether LLM-based digital twins can be reused as auxiliary measurements in behavioral research. As AI increasingly participates in the production of data and measurements, determining when AI-generated evidence can substitute for, supplement, or reduce human data collection becomes an important challenge.

LLMs are increasingly used to simulate survey respondents, consumers, political respondents, and experimental participants \citep{aher2023, argyle2023,santurkar2023,bisbee2024,cui2025}. These studies typically condition models on demographic, attitudinal, or persona information and evaluate their ability to reproduce human distributions, experimental effects, or held-out behaviors. As a result, LLM-based human simulation has become a growing research area across the social sciences, including within the information systems (IS) discipline (e.g., \citealt{gao2025}). The evidence, however, remains mixed. Synthetic respondents sometimes recover human means and treatment-effect directions, but they can also reduce variation, misrepresent subgroup patterns and conditional relationships, and overstate effect sizes \citep{santurkar2023,bisbee2024,boelaert2025,cui2025,ashokkumar2026}. Performance is also sensitive to design choices, including persona construction, answer presentation, and response generation \citep{dominguez2024,lutz2025,ahnert2026}. Consequently, success on a particular validation metric does not establish that a model can serve as a reliable substitute for human data, nor does it indicate how much human measurement can be reduced while preserving valid inference. This concern echoes a broader challenge in IS research on AI: predictive performance gains do not by themselves establish operational utility, making downstream consequences of model performance an important part of evaluation \citep{abbasi2024}.

Information-rich, person-specific \emph{digital twins} raise a more ambitious version of this problem. Instead of describing hypothetical respondents using short demographic or persona profiles, digital twin approaches collect extensive information about specific individuals and use it to predict their future responses. For example, \citet{park2024} construct agents for 1,052 people from qualitative interviews, while \citet{toubia2025} collect more than 500 demographic, psychological, economic, cognitive, and behavioral responses from 2,058 U.S. participants. These systems report strong held-out predictive performance and suggest a more ambitious possibility in that researchers may be able to reuse digital twins as substitutes or complements for future human data collection \citep{park2024,toubia2025}. Researchers could ask new questions or evaluate new scenarios without repeatedly collecting the same outcomes from the original participants. Yet, held-out accuracy and aggregate replication address only whether twins resemble humans. They do not establish how much human measurement can actually be reduced while preserving valid inference about human quantities of interest.

This gap motivates our focus on \emph{statistical substitutability}. We define this term as the extent to which predictions from a LLM-based digital twin pipeline can reduce the human measurement required for valid inference about a prespecified human estimand. Statistical substitutability differs from response accuracy, distributional similarity, and aggregate effect replication. A model may reproduce a human mean, distribution, or treatment effect while failing to predict which individuals contribute to that pattern \citep{tjuatja2024}. In such cases, the model may appear behaviorally realistic at the aggregate level while providing little information about the missing human outcomes needed for inference. Human validation data can correct average prediction error, but they cannot create respondent-level signal that is absent from the predictions \citep{angelopoulos2023,angelopoulos2024,broska2025}. For a treatment effect, a digital twin must predict who responds above or below the expectation for their condition, not merely reproduce the average difference between conditions. From an IS perspective, statistical substitutability provides a way to evaluate whether AI-generated observations can serve as reliable sources of evidence in empirical research rather than simply realistic simulations. More broadly, this concern aligns with emerging IS research that seeks systematic approaches for auditing and validating LLM-generated outputs in downstream applications \citep{lengyuan2026}.

This distinction matters when human measurement is incomplete. Planned and realized sample sizes can differ because of recruitment shortfalls or other deviations from the study plan \citep{claesen2021,nevins2022}. Digital twin predictions may then help close the remaining precision gap, but only if they contain sufficient respondent-level information about the missing human outcomes. This leads to three research questions. First, do LLM-based digital twins reproduce both aggregate human effects and the respondent-level signal required for statistical calibration? Second, can twin predictions compensate for a shortfall in human outcomes and recover the statistical precision of the reference human design? Third, do newer models, richer respondent representations, and additional human labels improve aggregate alignment and statistical substitutability in the same way? 

We answer these questions using established statistical calibration methods rather than treating twin responses as human observations. Prediction-powered inference (PPI) combines model predictions with a smaller human validation sample, using the observed outcomes to correct model error and the paired human-model association to improve precision \citep{angelopoulos2023,angelopoulos2024}. \citet{broska2025} connect this association to effective sample size and power in mixed human-AI designs. We build on this existing machinery rather than propose a new PPI estimator. Instead, we use it to develop an evaluation framework for statistical substitutability that assesses aggregate fidelity, paired respondent-level signal, finite-sample human-label recovery, and transport stability, allowing researchers to evaluate when AI-generated participants can meaningfully reduce human measurement while preserving valid inference.

We evaluate the framework in two complementary settings. The first uses Twin-2K \citep{toubia2025} to reconstruct 12 behavioral studies and compare aggregate human-twin replication with respondent-level signal and finite-sample prediction-assisted performance. Our findings do not contradict the original Twin-2K evidence that the twins achieve strong held-out predictive performance and reproduce some behavioral effects. Instead, they address a different question of whether those predictions contain enough respondent-level information to reduce future human measurement while preserving valid inference. This setting allows us to evaluate whether aggregate behavioral replication is accompanied by the respondent-level signal required for statistical substitutability.

The Moore-Berg extension tests one possible explanation for the weak Twin-2K signal. Many reconstructed Twin-2K tasks have no obvious relationship with the demographic or personality information used to represent respondents. We therefore add the partisan meta-perception study of \citet{mooreberg2020}, where party identity is directly related to the estimands, as a deliberately more favorable test of person-specific prediction. This setting also allows us to compare multiple models, respondent representations, and calibration strategies. The results provide a stronger test of whether improvements in model capability, respondent information, and human calibration translate into greater statistical substitutability. Because the Twin-2K target sample does not contain human Moore-Berg outcomes, the extension also highlights the challenges of establishing validity when calibration is carried across populations.

We make three contributions in this study. First, we introduce statistical substitutability as an estimand-specific criterion for evaluating when AI-generated participants can replace a portion of human measurement while preserving valid inference. This framing builds on mixed-subject work on PPI correlation, effective sample size, and power. Second, we develop a practical evaluation framework for statistical substitutability that integrates aggregate fidelity, paired respondent-level signal, finite-sample human-label recovery, and transport stability. Third, we provide empirical evidence that aggregate replication and inferential usefulness often diverge, demonstrating that digital twins can appear behaviorally plausible while providing little information for reducing future human measurement. Together, these contributions shift evaluation from behavioral resemblance to inferential usefulness, providing a framework for determining when AI-generated participants can serve as reliable sources of scientific evidence.

The remainder of the paper proceeds as follows. Section 2 reviews research on synthetic respondents, digital twins, and statistical calibration. Section 3 develops the methodological approach and empirical design. Section 4 reports the Twin-2K and Moore-Berg results. Finally, Section 5 discusses the implications of this work for generative AI-enabled behavioral research and, more generally, human-AI workflows.

\section{Related Work}

\subsection{Synthetic respondents and behavioral prediction}
Research on synthetic participants differs in how people are represented and how fidelity is evaluated. Persona-based studies condition models on demographic or attitudinal information. \citet{argyle2023} construct LLM samples (often referred to as silicon samples) from sociodemographic profiles, while \citet{santurkar2023} and \citet{bisbee2024} show that demographic or political conditioning can still leave important forms of misalignment. Synthetic averages may resemble human averages even when response variation, subgroup patterns, and regression relationships differ \citep{bisbee2024,boelaert2025}.

Other work asks whether LLM participants reproduce experiments or predict held-out behavior. Large-scale studies find that LLMs can recover some experimental findings and treatment-effect directions, but interactions are harder to reproduce and effect magnitudes can be distorted \citep{cui2025,ashokkumar2026}. Fine-tuning can improve prediction in unseen studies and conditions \citep{kolluri2025}. At the same time, direct human substitution remains risky. \citet{golisingh2024} find that LLM-elicited preferences can differ materially from human preferences even when the models remain useful for hypothesis generation, and \citet{gao2025} show that failures to reproduce human behavior vary across models and implementation choices. More broadly, recent IS research suggests that human-like AI performance must be evaluated across multiple dimensions rather than inferred from a single benchmark or behavioral similarity measure \citep{wangpeisun2026}.

No single metric captures all forms of fidelity. A simulation may recover selected means or effects while misrepresenting identity groups \citep{wangidentity2025}, compressing heterogeneity \citep{bisbee2024,boelaert2025}, or failing on more demanding experimental targets \citep{cui2025,ashokkumar2026}. Reviews therefore caution against treating synthetic samples as generally interchangeable with human data \citep{dillion2023,sarstedt2024}. This concern is particularly relevant for IS researchers, who commonly use surveys, experiments, and other human-centered methods to investigate topics such as technology use, digital work, platform participation, and human-AI interaction \citep{maruping2025quantitative}. 

\subsection{Behavioral fidelity, distributional alignment, and paired signal}

Much of the literature focuses on aggregate or distributional alignment. Common targets include marginal response distributions \citep{argyle2023,boelaert2025}, subgroup means and frequencies \citep{santurkar2023,bisbee2024}, regression relationships \citep{bisbee2024}, and treatment effects \citep{cui2025,ashokkumar2026}. Random silicon sampling is designed to reproduce group distributions without matching synthetic and human individuals \citep{sun2024}. Fine-tuning and supervision can also improve distributional alignment across surveys and populations \citep{cao2025,suh2025,wangfinetuning2025,kambhatla2026}. These are useful goals when the target is a population summary.

Population alignment and paired prediction are nevertheless different. A synthetic response vector could reproduce the human distribution and treatment effect after its values were randomly reassigned across people. The aggregate result would remain unchanged, but the predictions would contain no information about the corresponding human outcomes. The reverse can occur when predictions are systematically biased yet preserve individual ordering. Human labels can correct the common shift and still benefit from the paired association. Related work likewise shows that calibration can improve population summaries even when individual predictions remain imperfect \citep{krsteski2026}, and that plausible distributions do not necessarily imply human-like individual response patterns \citep{tjuatja2024}.

This distinction is critical for prediction-assisted inference. Aggregate agreement can be useful without supplying the respondent-level signal needed for human-data savings. For researchers considering AI-generated observations as substitutes or complements for human data collection, the latter is the more consequential property. We therefore measure paired human-twin association for the estimand itself and ask how much precision that signal can contribute under realistic human-label constraints.

\subsection{Information-rich digital twins}

Digital twins represent a specific observed person using a substantial response history. In \citet{park2024}, agents using interviews and surveys reproduce held-out General Social Survey answers at 86\% of participants' own test-retest consistency. Twin-2K-500 provides more than 500 responses per person and repeated behavioral tasks \citep{toubia2025}. Its baseline twins reach 71.7\% average predictive accuracy, about 87.7\% of the human test-retest benchmark, and reproduce about half of the evaluated behavioral effects. Related work on digital populations likewise seeks to reduce human recruitment costs while improving the fidelity and diversity of synthetic populations \citep{crowdllm2026}.

Richer conditioning and behavior-specific training can improve prediction in some settings. The representation of respondent information matters \citep{lutz2025}, fine-tuning on observed behavior can improve later-action prediction \citep{lu2025}, and specialized models such as Centaur improve prediction for held-out participants and new experiments \citep{binz2025}. These findings make information-rich digital twins a particularly favorable setting in which to test whether personalization creates inferential value.

Existing digital twin benchmarks mainly establish how well systems predict held-out responses or reproduce behavioral patterns. They do not establish how much future human measurement can actually be avoided for a downstream estimand. We address that gap by examining whether the person-specific information contained in digital twins is sufficiently strong and stable to produce statistical substitutability for downstream behavioral estimands.

\subsection{Statistical calibration}

A separate literature provides methods for combining predictions with observed outcomes without treating model output as human data. PPI and PPI++ use a labeled validation sample to correct prediction error and can improve precision when predictions carry useful signal \citep{angelopoulos2023,angelopoulos2024}. Related approaches include post-prediction correction \citep{wang2020}, design-based supervised learning \citep{egami2023}, and methods designed to protect efficiency when predictions are weak \citep{miao2025}. Work on machine-learning-generated variables reaches a similar conclusion. Prediction error can become measurement error in downstream inference and must be handled explicitly \citep{schecterli2025}.

\citet{broska2025} provide the closest precursor to our inferential framing. Their mixed-subject design treats LLM predictions as potentially informative observations while keeping human outcomes as the gold standard. Their framework therefore already shows how useful paired signal can translate into precision and human-sample savings. These methods establish how predictive signal can be translated into efficiency gains. They do not, however, provide a framework for evaluating whether a particular digital twin system contains enough signal to justify reducing human measurement in the first place. Our contribution is not another estimator, correlation measure, or allocation formula. Rather, we use these ideas to evaluate whether contemporary digital twins possess enough estimand-specific signal to justify reductions in human measurement. 

Recent applications extend the same general strategy. \citet{wangmarket2026} develop a consistent augmentation estimator for conjoint analysis and show that naive substitution can increase bias. \citet{krsteski2026} compare prompting, fine-tuning, and rectification under a fixed human-response budget. These methods establish how imperfect predictions can be used validly and when they can improve efficiency. What they leave open is the empirical question of whether current LLM-based digital twins satisfy those conditions reliably enough to justify reducing human measurement.

Our Moore-Berg extension adds a further boundary. Standard prediction-assisted designs are most straightforward when labeled and prediction-only observations represent the same target population. In our transported setting, a correction estimated among Moore-Berg respondents is applied to predictions for Twin-2K respondents. Generalizing an error relationship across samples requires additional assumptions about overlap and stability \citep{stuart2011,tipton2013}. We therefore treat transport stability as a separate part of the evaluation and describe transported results as agreement with the source reference rather than as validated target accuracy.

\subsection{From behavioral fidelity to statistical substitutability}
The four streams above answer different parts of the same problem. Research on using LLMs as human surrogates shows that models can reproduce some aggregate human patterns but that performance is uneven across outcomes and design choices. Distributional work shows that matching population summaries does not guarantee matched person-level variation. Digital twin research demonstrates that richer histories can improve person-specific prediction. Calibration and mixed-subject methods show how paired prediction signal, when present, can be converted into valid precision gains and lower human-sample requirements.

What remains unresolved for LLM-based digital twins is whether systems designed to represent specific individuals actually contain enough estimand-specific signal to deliver meaningful inferential gains in downstream studies. We address this question through four complementary evaluations. Specifically, we assess aggregate fidelity, paired respondent-level signal, finite-sample human-label recovery, and stability when calibration is carried across populations. This framework treats statistical substitutability as an empirical property of a digital twin pipeline for a particular estimand rather than as a general attribute of a model. Importantly, behavioral realism is not sufficient for statistical substitutability. For IS researchers, the central issue is not whether AI-generated observations look human, but whether they contain enough information to serve as reliable sources of data in empirical research workflows.

\section{Research Design and Methods}
\begin{figure}[!htbp]
    \centering
    \includegraphics[width=\linewidth]{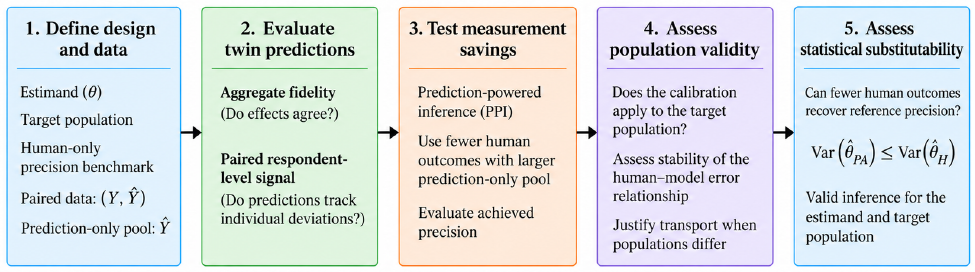}
    \caption{Workflow for evaluating statistical substitutability of
    digital twins. Here, $\widehat{\theta}_{\mathrm{PA}}$ is the
    prediction-assisted estimator using fewer observed human outcomes
    and additional twin predictions, while $\widehat{\theta}_H$ is
    the human-only estimator under the full reference design.
    Both estimate the same human estimand $\theta$.
    Statistical substitutability requires at least the reference
    precision while preserving valid inference in the target
    population.}
    \label{fig:substitutability-workflow}
\end{figure}
Figure~\ref{fig:substitutability-workflow} summarizes our proposed workflow for evaluating statistical substitutability. We begin by specifying the human estimand, target population, and human-only precision benchmark, then assemble paired human outcomes and twin predictions alongside a separate prediction-only pool. We evaluate aggregate fidelity and paired respondent-level signal separately because reproducing an
average effect does not establish that predictions track individual departures from the relevant group means. We then assess whether PPI can recover the benchmark precision with fewer human outcomes, accounting for limited validation data and a finite prediction-only pool. Population validity is a required component of the evaluation, with additional justification needed when calibration is carried across populations. Statistical substitutability is supported only when the reduced-human design preserves valid inference for the target estimand while attaining at least the reference precision. When population validity remains unresolved, precision gains alone cannot establish target substitutability.

\subsection{Statistical substitutability framework}

Our research design asks whether a digital twin pipeline can reduce human measurement for a particular estimand without sacrificing valid inference. The target is always a quantity about human outcomes. Twin responses enter as auxiliary predictions rather than as observed human data. We use established prediction-powered and mixed-subject methods to measure how much information those predictions contribute.

Let $\theta$ denote a human estimand and let $\widehat{\theta}_{H,N_0}$ denote the human-only estimator under a reference design with $N_0$ observed human outcomes. Suppose instead that only $n_L<N_0$ human outcomes are collected and twin predictions are available for those respondents and an additional prediction-only sample of size $N$. Let $\widehat{\theta}_{\mathrm{PA}}(n_L,N)$ denote a valid prediction-assisted estimator that uses the $n_L$ observed human outcomes together with those $N$ additional prediction-only observations. We call the digital twin pipeline \emph{statistically substitutable} for the omitted human measurement at label size $n_L$ when this prediction-assisted estimator can attain at least the precision of the reference human-only design using a feasible number of additional predictions.
\begin{equation}
\operatorname{Var}\!\left(\widehat{\theta}_{\mathrm{PA}}(n_L,N)\right)
\leq
\operatorname{Var}\!\left(\widehat{\theta}_{H,N_0}\right).
\label{eq:substitutability}
\end{equation}
The corresponding reduction in human measurement is $1-n_L/N_0$. This definition is \textit{estimand specific and design specific}. A model can be substitutable for one outcome or treatment effect and not for another, and the same model can become more or less useful as the human-label budget and available prediction-only pool change.

The evaluation begins with a human estimand and a reference level of precision. Digital twin predictions are paired with human outcomes in a validation subset. We first examine whether the twins reproduce the human quantity at the aggregate level and whether they predict respondent-level departures from the relevant group expectation. We next translate that paired signal into a best-case efficiency ceiling and then ask whether the same gain is achievable when the signal must be estimated from limited human labels and the prediction-only pool is finite. When calibration is carried from one population to another, stability of the human-model error relationship becomes an additional requirement.

This framework separates four questions that are often combined in digital twin validation. Aggregate fidelity asks whether twin-only results resemble the corresponding human result. Paired signal asks whether predictions contain information about the particular human outcomes that remain uncertain after accounting for the groups defining the estimand. Finite-sample feasibility asks whether the gain survives when the signal itself must be learned from limited labels and only a finite set of additional predictions is available. Transport stability is considered separately when a correction learned in one sample is applied to another. From an IS perspective, these evaluations determine whether AI-generated responses can serve as reliable inputs in behavioral research workflows.

\subsection{From twin signal to human-data savings}
We use established PPI results to connect paired human-twin signal to the criterion in Equation~\ref{eq:substitutability} \citep{angelopoulos2023,angelopoulos2024,broska2025}. PPI keeps observed human outcomes as the inferential baseline and allows twin predictions to improve precision only when they contain useful estimand-specific signal.

Let the prediction-only sample contain $N$ units with twin predictions $\widehat{Y}_1,\ldots,\widehat{Y}_N$. Let a separate validation sample contain $n$ units with both human outcomes $Y_1,\ldots,Y_n$ and corresponding predictions $\widehat{Y}_1,\ldots,\widehat{Y}_n$. For the population mean $\theta=E[Y]$, a power-tuned PPI estimator can be written as
\begin{equation}
\widehat{\theta}_{\lambda}
=
\underbrace{\frac{1}{N}\sum_{i=1}^{N}\lambda\widehat{Y}_i}_{\text{synthetic mean}}
+
\underbrace{\frac{1}{n}\sum_{j=1}^{n}\left(Y_j-\lambda\widehat{Y}_j\right)}_{\text{rectification}} .
\label{eq:ppi_mean}
\end{equation}
Here $\lambda\in[0,1]$ controls how strongly the prediction-based adjustment enters the estimator. When $\lambda=0$, Equation~\ref{eq:ppi_mean} reduces to the human-only mean. When $\lambda=1$, the predictions receive full weight and the validation sample corrects their average error. PPI++ chooses $\lambda$ to reduce sampling variance \citep{angelopoulos2023,angelopoulos2024}. We use the power-tuned notation adopted by \citet{krsteski2026}. The correction preserves the human estimand even when the raw twin mean is biased, but it improves precision only when the predictions contain useful paired signal.

The key implication is that the value of a digital twin depends primarily on its paired association with human outcomes rather than its aggregate fidelity or predictive accuracy. Appendix~\ref{sec:appendix-ppi-details} provides the variance decomposition and coefficient details. For a population mean, let $\rho=\operatorname{Corr}(Y,\widehat{Y})$ and let $\kappa=n/N$. The unconstrained variance-minimizing coefficient implies the following idealized variance-reduction ceiling:
\begin{equation}
\operatorname{VR}_{\max}
=
\frac{\rho^2}{1+\kappa}.
\label{eq:ppi_vrmax}
\end{equation}
As the prediction-only sample becomes large relative to the human sample, $\kappa$ approaches zero and the ceiling approaches $\rho^2$. This is a best-case calculation rather than a guarantee of achievable savings. It treats the paired association as known and assumes that enough prediction-only observations are available. If the unconstrained optimum falls outside $[0,1]$, the implemented procedure uses the nearest boundary. The choice $\lambda=0$ returns the human-only estimator, so the population-level optimized procedure cannot be less precise than human-only estimation. This guarantee does not apply to a fixed unit coefficient, which can increase variance when the predictions are poorly scaled. Appendix~\ref{sec:appendix-lambda} gives that case explicitly.

We express variance reduction in human-sample terms using the effective-sample-size multiplier $1/(1-\operatorname{VR})$. For example, a multiplier of 1.072 means that the prediction-assisted estimate has the same variance as a human-only estimate with about 7.2\% more labeled observations. A large synthetic sample therefore matters only through the information it contributes about the human estimand.

For a two-condition effect, the useful signal is not whether human and twin outcomes both differ across treatment conditions. That shared difference can be generated by the condition itself. A twin helps replace missing human outcomes only if it also predicts which respondents are unusually high or low relative to others in the same condition. We therefore use the residualized correlation
\begin{equation}
\rho_T
=
\operatorname{Corr}\!\left(
Y-E[Y\mid T],
\widehat{Y}-E[\widehat{Y}\mid T]
\right).
\label{eq:rho}
\end{equation}
This quantity removes the condition means before measuring human-twin association. A twin can reproduce an aggregate treatment effect while having $\rho_T$ near zero if it does not predict who departs from that condition mean. In the large prediction-only-sample linear approximation, $\rho_T^2$ is the idealized efficiency ceiling. Exact treatment-effect variance also depends on the arm-specific variances and covariances \citep{broska2025}.

The theoretical ceiling asks how useful the twin could be if its paired signal were known and enough predictions were available. The finite-sample benchmark asks whether a limited human validation sample can estimate enough signal to recover the precision of a larger human-only study using the predictions that are actually available. Following \citet{broska2025}, let $n_L$ be the labeled human sample, let $N_0$ be the reference human-only size, and let $\rho_+$ be the nonnegative estimated residualized correlation. A finite prediction-only requirement exists only when
\begin{equation}
\rho_+^2>1-\frac{n_L}{N_0}.
\label{eq:threshold}
\end{equation}
A small positive correlation may imply some asymptotic efficiency gain while still being too weak to replace a prespecified share of the human sample. When the threshold is satisfied, we calculate the required prediction-only size using Equation~\ref{eq:nreq} in Appendix~\ref{sec:appendix-ppi-details}. The design is feasible only if that requirement also fits within the available disjoint prediction pool. We therefore report both the best-case efficiency implied by full-data paired signal and the human-label savings achieved in repeated finite samples.

\subsection{Empirical evaluation}

We evaluate the framework in two complementary settings. Twin-2K is the primary broad benchmark because it contains information-rich person-specific profiles, matched human outcomes and twin predictions, and multiple behavioral tasks. It allows us to ask whether the strong predictive performance reported for digital twins translates into paired signal and human-data savings across different estimands. The Moore-Berg extension is deliberately more favorable to personalization because party identity is directly related to the estimands. It also allows us to compare newer model generations, alternative respondent representations, and the additional difficulty created when calibration is carried from a source sample to a different target population.

Across both settings, model responses are treated as predictions rather than human observations. The Twin-2K benchmark uses the released baseline predictions rather than regenerating responses. For the Moore-Berg extension, we use model-specific prompts and hold generation, parsing, and respondent-level deduplication fixed within each source and target regime. We report all six models and all three target persona regimes. We do not choose a preferred specification from the final correlation or variance-reduction results. The estimates therefore describe the evaluated pipelines rather than the best performance that any possible digital twin procedure could attain. Appendix~\ref{sec:oa-reproducibility} documents the resampling and validation procedures, and interpretation boundaries. Appendix~\ref{sec:appendix-genai_doc} documents GenAI use, including model use, prompting, generation, validation, and human
verification.

\subsubsection{Twin-2K benchmark}

Twin-2K-500 contains 2,058 U.S. participants who completed four survey waves and more than 500 questions \citep{toubia2025}. The released baseline simulations use GPT-4.1-mini conditioned on a text persona that excludes the held-out task response used for evaluation. We reconstruct 12 two-condition contrasts with treatment indicator $T$, human outcome $Y$, and twin prediction $\widehat{Y}$. Specifically, the tasks cover Outcome Bias, Sunk Cost, Less-is-More, Framing, Absolute versus Relative Savings, Myside Bias, Anchoring, Proportion Dominance, WTA/WTP, Allais, Linda, and Base Rate. Table~\ref{tab:appendix-scope} in Appendix~\ref{sec:appendix-scope} records the exact contrasts and exclusions.

For each task, the human and twin effects are
\begin{align}
\tau_H &= E[Y\mid T=1]-E[Y\mid T=0], \\
\tau_D &= E[\widehat{Y}\mid T=1]-E[\widehat{Y}\mid T=0].
\end{align}
Here $T=1$ and $T=0$ denote the two assigned conditions. Aggregate fidelity is assessed by comparing the human contrast $\tau_H$ with its digital-twin counterpart $\tau_D$. We report the effects in original units and standardize them by the root mean of the two within-condition variances. The comparison of $\tau_H$ and $\tau_D$ measures aggregate fidelity. Equation~\ref{eq:rho} measures the paired respondent-level signal that determines the efficiency ceiling. The finite-sample benchmark then asks whether that signal is strong enough to satisfy Equation~\ref{eq:substitutability} under reduced human-label budgets.

\paragraph{Finite-sample benchmark.}
Human studies are commonly planned around a target sample size, but the realized sample can fall short because of under-recruitment or other deviations from the study plan \citep{claesen2021,nevins2022}. We therefore interpret the 90\% condition as a last-mile shortfall scenario. Most of the intended human outcomes have been obtained, and the question is whether twin predictions for additional eligible respondents can recover the precision of the original human-only design. This complements the prospective planning problem studied by \citet{broska2025}. Rather than choosing an optimal human-prediction mix in advance, we fix the realized human fraction and ask whether the available predictions can make up the missing precision. The 25\%, 50\%, and 75\% conditions provide progressively more demanding stress tests.

We mask human outcomes retrospectively and evaluate three reference designs. The first uses Twin-2K's realized sample size. The other two use balanced designs powered at 80\% for Cohen's $d=0.2$ or $d=0.5$ with two-sided $\alpha=0.05$. Human-label fractions are 0.25, 0.50, 0.75, and 0.90. Each design uses 500 repetitions.

Within each repetition, we draw a balanced labeled sample and remove those respondent identifiers from the prediction-only pool. We estimate the residualized correlation from the labeled observations and use Equations \ref{eq:threshold} and \ref{eq:nreq} to determine whether the omitted human precision can in principle be recovered. We run prediction-powered OLS for the treatment coefficient only when the required prediction-only sample is finite and fits in the remaining disjoint pool. Achieved variance reduction compares PPI and human-only estimates over the same successful repetitions. This repeated-sample step distinguishes the efficiency ceiling implied by full-data paired signal from the human-label savings that are actually attainable when the signal itself must be estimated.

\subsubsection{Moore-Berg extension}

The Moore-Berg setting serves three purposes in the framework. First, it provides an estimand for which a core respondent attribute, party identification, is directly relevant to the outcome contrasts. This creates a more favorable test of whether personalization produces useful paired signal than many of the reconstructed Twin-2K tasks. Second, it allows us to compare six model versions and three respondent representations. Third, it creates a source-to-target setting in which we can examine how calibration behaves when the human-model relationship estimated in one sample is carried to another.

\citet{mooreberg2020} examine whether American Democrats and Republicans accurately understand how the other party views them. In Study 1, a nationally representative sample of 1,056 partisans rated their own warmth and humanity toward Democrats and Republicans. They also estimated how the opposing party would make the same ratings. Actual prejudice is the difference between warmth toward one's own party and warmth toward the opposing party. Actual dehumanization is the corresponding difference in ratings of how evolved and civilized the two parties are. Meta-prejudice and meta-dehumanization use the same contrasts but are based on what respondents believe the opposing party would report. \citet{mooreberg2020} find that Democrats and Republicans express similar levels of out-party prejudice and dehumanization, yet each side believes the other is roughly twice as negative as it reports. Our extension focuses on the accuracy of these meta-perceptions rather than the downstream outcomes studied in the original paper.

We reconstruct four survey-weighted cross-party gaps. Democratic meta-prejudice minus Republican actual prejudice measures how much Democrats overestimate Republican affective prejudice toward Democrats. Democratic meta-dehumanization minus Republican actual dehumanization measures the analogous gap in humanity ratings. The other two estimands reverse the party roles. These are not simple comparisons of whether Democrats or Republicans are more prejudiced. They measure how much each side exaggerates the other side's reported attitudes. Positive values mean overestimation. The published human gaps are 23.95, 33.20, 25.43, and 37.62, and our reconstructed values match them.

We generate source predictions for the Moore-Berg respondents and target predictions for 1,387 Democratic or Republican Twin-2K respondents. The model set includes GPT-5.4, GPT-4.1-mini, Qwen3.5-27B, Qwen3.8-27B, Llama 4 Scout, and Llama 3.1 8B. The party-only target prompt supplies only party affiliation. The common-fields prompt supplies a structured profile. Nine fields are available in both the Moore-Berg and Twin-2K samples, namely party, ideology, age, gender, race or ethnicity, education, household income, U.S. region, and employment status. The persona-summary prompt uses a free-text summary of the broader Twin-2K profile. Appendix~\ref{sec:appendix-moore-berg} summarizes the estimands and prompt regimes.

For descriptive within-family comparisons, we pair GPT-4.1-mini with GPT-5.4, Llama 3.1 8B with Llama 4 Scout, and Qwen3.5-27B with Qwen3.8-27B. These are comparisons between evaluated model versions, not controlled tests of parameter scaling. The pairs differ in architecture, training, and other features.

For each model, source paired signal is the survey-weighted version of Equation~\ref{eq:rho}, calculated within the relevant party and outcome components. We summarize source replication by the mean absolute deviation between model and human contrasts. Human-label budgets are 25, 50, 100, 115, 250, and 500, with 500 repetitions. These analyses ask whether model generation and respondent representation change both aggregate agreement and the paired signal that drives statistical substitutability.

The source-to-target analysis adds a separate transport question. Here $S$ denotes the Moore-Berg source sample and $T$ denotes the Twin-2K target sample. The quantity $\widehat{\tau}_{Y,S}$ is the human contrast estimated from the labeled Moore-Berg respondents, $\widehat{\tau}_{\widehat{Y},S}$ is the corresponding model-predicted contrast in the source sample, and $\widehat{\tau}_{\widehat{Y},T}$ is the model-predicted target contrast. We estimate the additive source correction
\begin{equation}
\widehat{\Delta}_S
=
\widehat{\tau}_{Y,S}-\widehat{\tau}_{\widehat{Y},S}
\end{equation}
and apply it to the target prediction
\begin{equation}
\widehat{\tau}_{T,\mathrm{cal}}
=
\widehat{\tau}_{\widehat{Y},T}+\widehat{\Delta}_S.
\label{eq:transportcal}
\end{equation}
Thus, $\widehat{\Delta}_S$ is the estimated source prediction error for the source contrast, and $\widehat{\tau}_{T,\mathrm{cal}}$ is the target model contrast after adding that source correction. Moore-Berg survey weights are used, but the source sample is not reweighted toward Twin-2K.

We also estimate
\begin{equation}
\widehat{\tau}_T(\lambda)
=
\widehat{\tau}_{Y,S}
+
\lambda\left(
\widehat{\tau}_{\widehat{Y},T}
-
\widehat{\tau}_{\widehat{Y},S}
\right).
\label{eq:lambda}
\end{equation}
We compare a unit coefficient, an adaptive coefficient estimated from each labeled subset, and a coefficient estimated from the complete source sample. Because the Twin-2K target sample does not contain human responses to the Moore-Berg outcomes, these transported estimates cannot establish target statistical substitutability or target accuracy. They instead test how sensitive the source correction is to limited labels and whether the source-to-target prediction gap magnifies instability when calibration is moved across populations. Appendix~\ref{sec:appendix-lambda} derives the coefficients, and Section~\ref{sec:oa-adaptive-transport} explains how estimation error is amplified under transport.

\section{Results}

The results evaluate statistical substitutability in stages. We first ask whether aggregate behavioral replication is accompanied by the paired respondent-level signal needed for prediction-assisted inference. We then ask whether that signal is strong enough to recover the precision lost when human measurement is reduced. The Moore-Berg extension provides a more favorable test of person-specific signal and asks whether newer models, richer respondent representations, and additional human labels improve the dimensions of performance that matter for statistical substitutability.

\subsection{Aggregate fidelity does not imply statistical substitutability}

Aggregate replication and respondent-level surrogate signal often diverge in Twin-2K. Table~\ref{tab:twin-results} shows this most clearly for Framing, where the human standardized effect is 0.761 and the twin effect is 0.941, but the residualized correlation is only 0.003. Framing illustrates the paper's central distinction. By an aggregate replication criterion, the twin performs well. By the criterion relevant for reducing future human measurement, it contributes essentially no paired information about which respondents depart from their condition mean. Richer participant histories may make predictions more person-specific on average, but the remaining personalization is weak and outcome dependent.

Strong respondent-level signal also does not guarantee correct aggregate replication, and matching the effect direction does not guarantee matching its magnitude. Proportion Dominance has the largest residualized correlation at 0.260 but human and twin effects with opposite signs, while Anchoring has a positive twin effect of 0.119 compared with 0.925 for humans. These cases show that aggregate direction, effect magnitude, and respondent-level signal must be evaluated separately. Neither aggregate fidelity nor paired signal can be inferred reliably from the other.

\begin{table}[!htbp]
\centering
\small
\begin{threeparttable}
\caption{Twin-2K aggregate effects and individual surrogate signal}
\label{tab:twin-results}
\begin{tabular}{lrrrr}
\toprule
Study & Human $d$ & Twin $d$ & Residualized $\rho$ & Max. VR \\
\midrule
Proportion Dominance & 0.224 & -0.040 & 0.260 & 6.74\% \\
Myside Bias & 0.344 & 0.175 & 0.170 & 2.90\% \\
Less-is-More & 0.739 & -0.243 & 0.078 & 0.61\% \\
WTA/WTP & 1.682 & 1.351 & 0.072 & 0.52\% \\
Base Rate & 0.851 & 3.846 & 0.033 & 0.11\% \\
Linda / Conjunction & 0.786 & 1.884 & 0.031 & 0.10\% \\
Outcome Bias & 0.493 & -0.017 & 0.025 & 0.06\% \\
Absolute / Relative Savings & 0.967 & 2.223 & 0.009 & 0.01\% \\
Framing & 0.761 & 0.941 & 0.003 & 0.00\% \\
Allais & 0.534 & \NA\tnote{a} & \NA & 0.00\% \\
Sunk Cost & -0.620 & 0.908 & -0.000 & 0.00\% \\
Anchoring & 0.925 & 0.119 & -0.008 & 0.01\% \\
\bottomrule
\end{tabular}
\begin{tablenotes}[flushleft]\footnotesize
\item Notes. $d$ standardizes the condition contrast by within-condition variability. Max. VR is $\rho^2$, an idealized large prediction-only-sample ceiling rather than achieved PPI performance.
\item[a] Allais twin predictions are constant within both conditions and have an aggregate contrast of zero. The standardized twin effect and residualized correlation are therefore undefined. Max. VR is zero because the predictions contain no usable variation.
\end{tablenotes}
\end{threeparttable}
\end{table}

\begin{figure}[!htbp]
\centering
\includegraphics[width=0.88\linewidth]{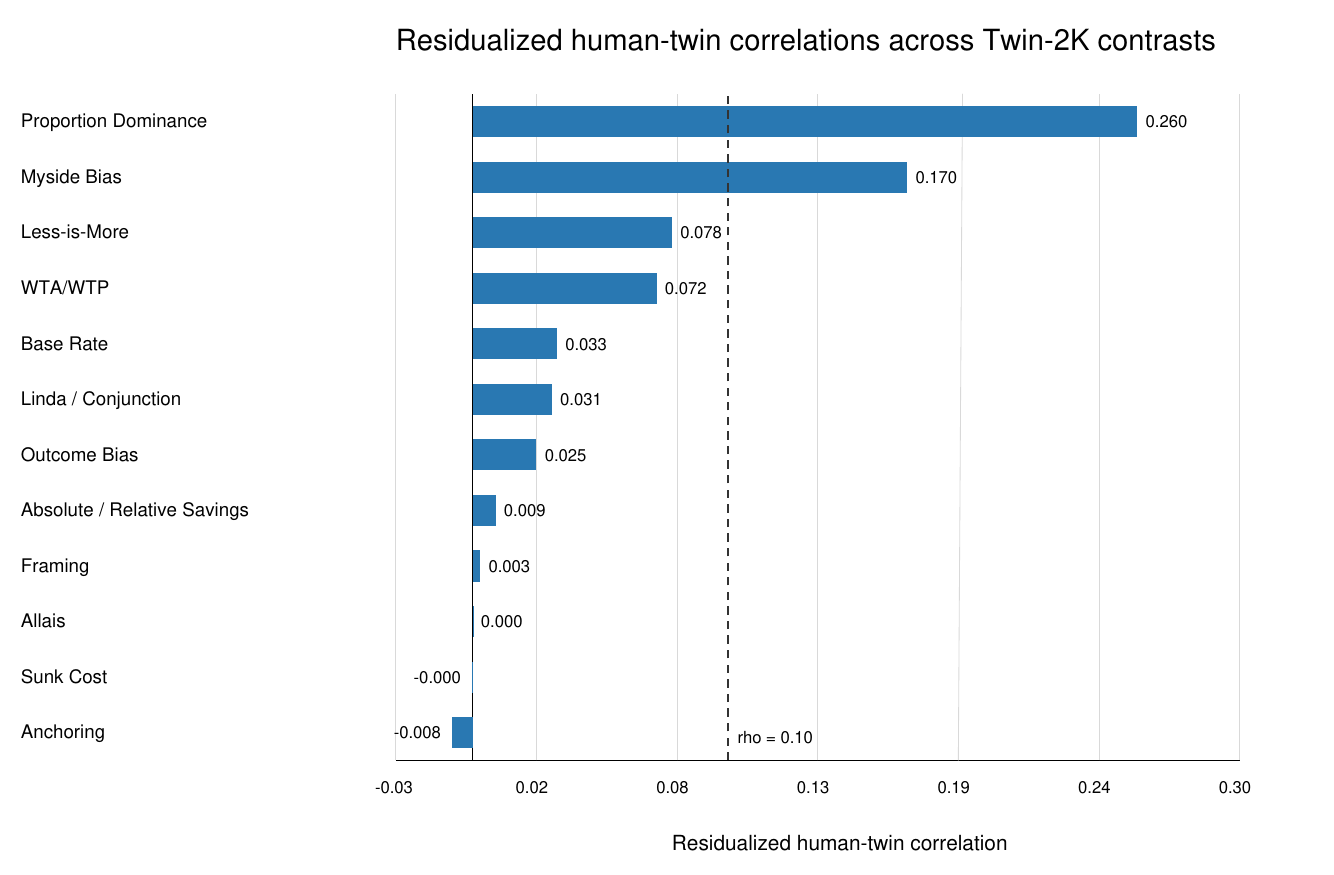}
\caption{Residualized human-twin correlations across the 12 primary Twin-2K contrasts. The dashed reference marks $\rho=0.10$, which corresponds to a 1\% idealized variance-reduction ceiling.}
\label{fig:twin-rho}
\end{figure}

Most Twin-2K contrasts contain very little respondent-level signal even when their aggregate effects look plausible. This finding is notable because Twin-2K represents one of the most information-rich digital-twin datasets currently available. Figure~\ref{fig:twin-rho} shows a median residualized correlation of 0.028, with 10 of 12 contrasts below 0.10 in absolute value and only two above that threshold. We use $|\rho|=0.10$ as a practical benchmark because $\rho^2$ is the idealized large-prediction-sample variance-reduction ceiling, so a correlation below 0.10 in absolute value implies less than 1\% maximum variance reduction. Even the strongest task has an idealized maximum variance reduction of 6.7\%, equivalent to an ESS multiplier of 1.072. Myside Bias reaches 2.9\%, while every remaining contrast is below 0.7\%.

Taken together, the Twin-2K results answer the first part of the framework negatively for most tasks. Aggregate behavioral replication can occur without the paired respondent-level information needed to reduce uncertainty about the corresponding human estimand.

\subsection{From theoretical signal to feasible human-data savings}

Finite-sample PPI feasibility is more demanding than the theoretical variance-reduction ceilings suggest. At the 90\% human-label fraction, the required labeled-sample correlation is approximately 0.316 to 0.318 across the realized study sizes, while the largest complete-data correlation is only 0.260. The theoretical ceilings also assume an arbitrarily large prediction-only sample and treat the correlation as known, whereas the operational benchmark must estimate the correlation from a smaller labeled subset. We emphasize the 90\% condition because it is the least aggressive reduction in human measurement and has the lowest required-correlation threshold. The 25\%, 50\%, and 75\% conditions also produced no successful repetitions under the realized-study reference design.

The Twin-2K predictions cannot rescue a study that is 10\% short of its target human sample. Across all 12 contrasts and 500 repetitions, the realized-study design produces no finite prediction-only requirement at the 90\% label fraction. The 90\% benchmark represents a study that has collected only 90\% of its reference human sample and asks whether twin predictions can provide enough information to make up the missing 10\%. Even under the favorable infinite-prediction ceiling, the strongest observed twin signal could reduce the human requirement only to about 93.3\%. A study at 90\% is therefore already below the level that these twins can recover.

Reducing the reference study size makes PPI occasionally feasible, but success remains rare. In the $d=0.5$ sensitivity design with $N_0=128$ and $n_L=114$, Proportion Dominance succeeds in 17.8\% of repetitions and yields 2.9\% conditional variance reduction across 89 successful repetitions, while Myside Bias succeeds in only 3.0\%. A repetition is successful only when the labeled-sample correlation implies a finite prediction-only requirement, the remaining disjoint Twin-2K pool contains enough balanced prediction-only observations, and the PPI estimate and confidence interval are computed successfully. The remaining tasks succeed in at most 1.0\% of repetitions. Table~\ref{tab:appendix-sensitivity} in Appendix~\ref{sec:oa-sensitivity} reports the full diagnostics. The main conclusion is therefore not that PPI is invalid, but that the released Twin-2K predictions provide too little within-condition signal for reliable human-label savings. The finite-sample results also show why a nonzero theoretical efficiency ceiling should not be interpreted as evidence of statistical substitutability on its own.

\subsection{A favorable Moore-Berg test of person-specific signal}

The Twin-2K results could partly reflect a mismatch between the information contained in the twins and the outcomes being predicted. Moore-Berg provides a more favorable test because party identification is directly related to the estimands being predicted. If estimand-relevant respondent information and stronger models are sufficient to produce statistical substitutability, this setting should show clearer gains in paired signal and inferential efficiency than the broad Twin-2K benchmark.

Later model versions improve aggregate source replication in all three model families, but they are not uniformly better on respondent-level signal and their potential efficiency gains remain modest. Table~\ref{tab:family-comparison} shows that mean absolute source deviation falls from 44.8 to 23.3 for GPT, from 80.4 to 69.4 for Llama, and from 18.0 to 15.8 for Qwen, while mean residualized correlation rises for GPT and Llama but falls slightly from 0.195 to 0.187 for Qwen. Oracle fixed-coefficient variance reduction rises in all three families but reaches only 6.9\% at its highest value. The more favorable setting therefore produces somewhat stronger respondent-level signal than most Twin-2K tasks, but improvement in one dimension still does not imply improvement in another.

\begin{table}[!htbp]
\centering
\small
\begin{threeparttable}
\caption{Descriptive within-family model-generation comparisons}
\label{tab:family-comparison}
\begin{tabular}{lllll}
\toprule
Family & Earlier to later model & Source deviation & Mean $\rho_S$ & Oracle fixed VR \\
\midrule
GPT & GPT-4.1-mini $\rightarrow$ GPT-5.4 & 44.8 $\rightarrow$ 23.3 & 0.190 $\rightarrow$ 0.222 & 4.1\% $\rightarrow$ 6.0\% \\
Llama & Llama 3.1 8B $\rightarrow$ Llama 4 Scout & 80.4 $\rightarrow$ 69.4 & 0.017 $\rightarrow$ 0.111 & 0.4\% $\rightarrow$ 2.2\% \\
Qwen & Qwen3.5-27B $\rightarrow$ Qwen3.8-27B & 18.0 $\rightarrow$ 15.8 & 0.195 $\rightarrow$ 0.187 & 5.6\% $\rightarrow$ 6.9\% \\
\bottomrule
\end{tabular}
\begin{tablenotes}[flushleft]\footnotesize
\item Notes. Lower source deviation is better. The comparisons are descriptive version contrasts rather than causal parameter-scaling tests. Oracle fixed VR uses the common-fields target regime and a coefficient estimated from the complete Moore-Berg source sample, then held fixed across the repeated smaller labeled samples. It is an oracle only for the low-label coefficient-estimation problem. It does not use the unobserved Twin-2K human target outcomes and is not an oracle for target accuracy.
\end{tablenotes}
\end{threeparttable}
\end{table}

\begin{figure}[!htbp]
\centering
\includegraphics[width=0.82\linewidth]{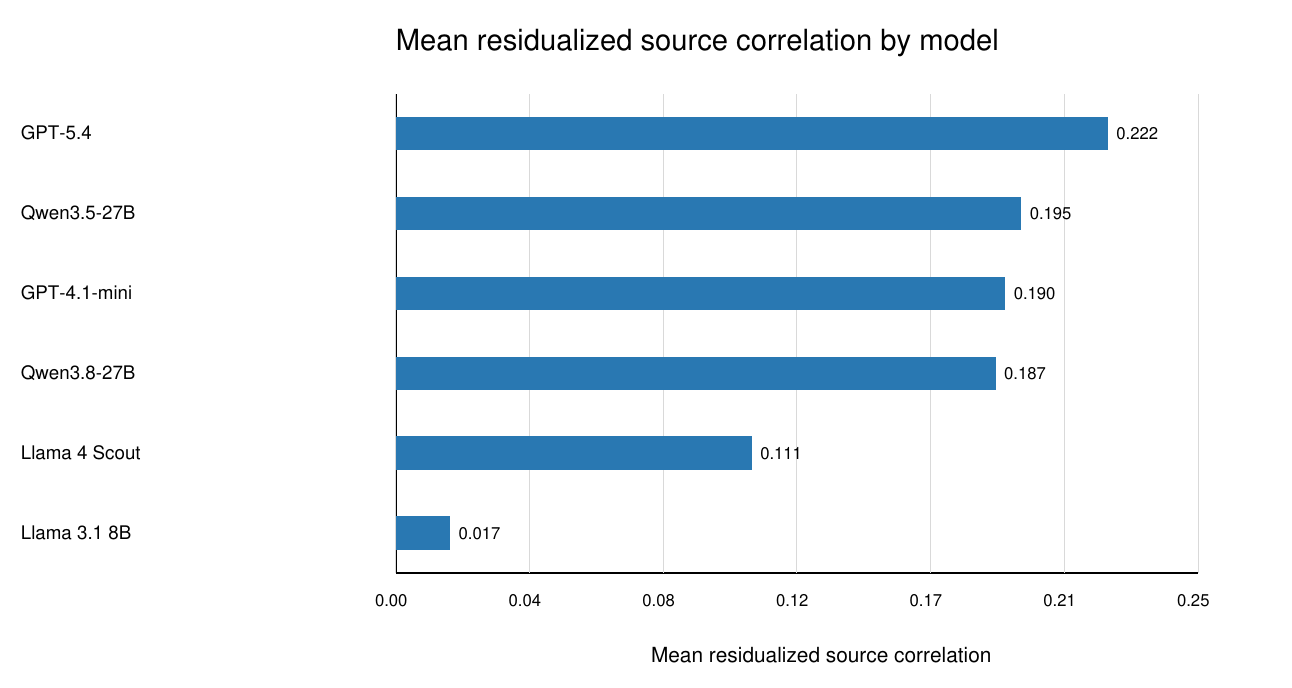}
\caption{Mean survey-weighted residualized source correlation across the four Moore-Berg estimands.}
\label{fig:model-rho}
\end{figure}

Aggregate fidelity and respondent-level signal produce different rankings across the six models. Figure~\ref{fig:model-rho} shows GPT-5.4 with the highest mean residualized correlation at 0.222, while Table~\ref{tab:family-comparison} shows that the lowest mean absolute source deviations belong to Qwen3.8 at 15.8 and Qwen3.5 at 18.0. GPT-4.1-mini and Qwen3.8 still have correlations near 0.19 despite very different aggregate deviations. Llama 4 Scout reaches 0.111, while Llama 3.1 8B is effectively uninformative at 0.017. The important comparison is therefore not which model ranks first on a single metric. Model quality depends on the downstream use. A model that better matches aggregate human results need not be better at predicting the respondent-level variation that drives inferential efficiency. Table~\ref{tab:appendix-rho} in Appendix~\ref{sec:appendix-moore-berg} reports the corresponding source correlations separately for each estimand.

Variance reduction matters because it is the link between predictive signal and human-data savings. Table~\ref{tab:family-comparison} shows that the later model has higher oracle fixed VR in every family, but the best value is only 6.9\%. Positive VR means the prediction-assisted estimator is less noisy than the human-only estimator at the same label budget, which translates into a larger effective sample size or fewer human observations for the same precision. Zero VR means no efficiency gain, and negative VR means the augmentation makes the estimate less precise. The broader result is that improvements in model generation do not consistently translate into proportional improvements in inferential usefulness. Later models are better on several dimensions, but the gains remain modest for statistical substitutability.

\subsection{Calibration corrects aggregate error more reliably than it improves precision}

Human labels substantially reduce source-reference deviation for most models, and the amount of correction depends on the label budget. This analysis separates two roles that human validation data can play. Labels can correct systematic model error, but statistical substitutability additionally requires the predictions to contain paired information that improves precision. Table~\ref{tab:calibration-budget} reports the raw deviation and the calibrated deviation across six fixed human-label budgets in the common-fields regime. Panel A reports the deviation that remains after calibration. Panel B reports the reduction from the raw starting point, calculated as raw deviation minus the calibrated deviation in Panel A. Since Panel B depends on each model's starting error, a large reduction does not necessarily imply the best calibrated result.

\begin{table}[!htbp]
\centering
\scriptsize
\caption{Moore-Berg common-fields source-reference calibration across fixed human-label budgets}
\label{tab:calibration-budget}
\begin{tabular}{lrrrrrrr}
\toprule
Model & Raw & 25 & 50 & 100 & 115 & 250 & 500 \\
\midrule
\multicolumn{8}{l}{\textit{Panel A. Mean calibrated source-reference deviation}} \\
GPT-5.4 & 26.47 & 14.52 & 10.89 & 7.73 & 7.15 & 4.97 & 3.73 \\
GPT-4.1-mini & 46.63 & 14.99 & 11.14 & 7.85 & 7.29 & 4.91 & 3.48 \\
Qwen3.5-27B & 21.78 & 15.56 & 11.79 & 8.85 & 8.27 & 6.49 & 5.44 \\
Qwen3.8-27B & 20.48 & 15.96 & 12.07 & 9.02 & 8.45 & 6.65 & 5.66 \\
Llama 4 Scout & 39.90 & 31.81 & 30.29 & 29.83 & 29.77 & 29.44 & 29.49 \\
Llama 3.1 8B & 75.46 & 18.17 & 13.58 & 10.63 & 9.87 & 7.54 & 6.56 \\
\addlinespace
\multicolumn{8}{l}{\textit{Panel B. Reduction from raw source-reference deviation}} \\
GPT-5.4 & \NA & 11.95 & 15.59 & 18.74 & 19.32 & 21.51 & 22.74 \\
GPT-4.1-mini & \NA & 31.64 & 35.48 & 38.77 & 39.34 & 41.72 & 43.15 \\
Qwen3.5-27B & \NA & 6.22 & 9.99 & 12.92 & 13.51 & 15.28 & 16.34 \\
Qwen3.8-27B & \NA & 4.52 & 8.41 & 11.45 & 12.03 & 13.83 & 14.82 \\
Llama 4 Scout & \NA & 8.10 & 9.62 & 10.07 & 10.13 & 10.47 & 10.41 \\
Llama 3.1 8B & \NA & 57.30 & 61.89 & 64.84 & 65.59 & 67.92 & 68.91 \\
\bottomrule
\end{tabular}
\begin{minipage}{0.98\linewidth}\footnotesize
Entries average across the four Moore-Berg estimands and 500 repetitions at each fixed human-label budget in the common-fields regime. Lower calibrated deviation is better. Reduction equals raw deviation minus calibrated deviation. These quantities measure alignment with the Moore-Berg human source reference, not error against an observed Twin-2K human target effect.
\end{minipage}
\end{table}

Calibration improves as more labels are added for five of the six models, while Llama 4 Scout largely plateaus after the smallest budget. At 500 labels, GPT-5.4 and GPT-4.1-mini reach calibrated deviations below 4, Qwen3.5 and Qwen3.8 fall to 5.44 and 5.66, and Llama 3.1 8B falls from a raw deviation of 75.46 to 6.56. Llama 4 Scout behaves differently. Most of its improvement occurs by 25 labels, after which calibrated deviation remains near 29-30. The pattern suggests that additional human labels can correct large systematic offsets when the source error is learnable, but more labels cannot compensate for every form of model mismatch.

Better aggregate calibration does not necessarily produce better inferential efficiency. Tables~\ref{tab:model-efficiency} and~\ref{tab:moore-budget} in Appendix~\ref{sec:appendix-lambda} show modest positive oracle variance reductions for the stronger models but often negative variance reduction for the feasible adaptive estimator. The first table reports the $n_L=115$ comparison, and the second reports all six human-label budgets. The calibration results therefore reinforce the distinction between correction and substitution. Human labels can remove substantial aggregate bias even when the model contains too little stable paired signal for those predictions to reduce sampling uncertainty. In other words, calibration can improve behavioral fidelity without improving statistical substitutability.

\subsection{Richer personas do not guarantee greater substitutability}

Richer persona representations do not consistently improve source-reference alignment. Table~\ref{tab:prompt-calibration} shows that Qwen3.8 has raw absolute deviations of 20.48 with common fields, 23.77 with party only, and 29.23 with a persona summary, while Qwen3.5 has nearly identical party-only and common-fields deviations of 21.70 and 21.78 but a worse persona-summary deviation of 23.86. GPT-5.4 follows a different ordering, with its persona summary performing best before calibration. The relative performance of the prompt regimes therefore depends on the model rather than on a simple ranking by persona richness.

\begin{table}[!htbp]
\centering
\scriptsize
\caption{Raw and calibrated deviation from the Moore-Berg source reference at the largest fraction-based label setting}
\label{tab:prompt-calibration}
\begin{tabular}{llrrr}
\toprule
Model & Prompt & Raw deviation & Calibrated deviation & Reduction \\
\midrule
GPT-5.4 & Common fields & 26.47 & 7.15 & 19.32 \\
GPT-5.4 & Persona summary & 23.68 & 7.37 & 16.31 \\
GPT-5.4 & Party only & 26.09 & 7.38 & 18.71 \\
GPT-4.1-mini & Common fields & 46.63 & 7.29 & 39.34 \\
GPT-4.1-mini & Party only & 48.45 & 8.16 & 40.30 \\
GPT-4.1-mini & Persona summary & 47.08 & 8.46 & 38.62 \\
Qwen3.5-27B & Party only & 21.70 & 7.77 & 13.93 \\
Qwen3.5-27B & Common fields & 21.78 & 8.27 & 13.51 \\
Qwen3.5-27B & Persona summary & 23.86 & 8.51 & 15.36 \\
Qwen3.8-27B & Common fields & 20.48 & 8.45 & 12.03 \\
Qwen3.8-27B & Party only & 23.77 & 9.67 & 14.10 \\
Qwen3.8-27B & Persona summary & 29.23 & 14.61 & 14.61 \\
Llama 3.1 8B & Common fields & 75.46 & 9.87 & 65.59 \\
Llama 3.1 8B & Persona summary & 74.86 & 14.14 & 60.72 \\
Llama 3.1 8B & Party only & 65.09 & 16.61 & 48.48 \\
Llama 4 Scout & Common fields & 39.90 & 29.77 & 10.13 \\
Llama 4 Scout & Persona summary & 36.95 & 32.48 & 4.47 \\
Llama 4 Scout & Party only & 32.92 & 36.55 & -3.63 \\
\bottomrule
\end{tabular}
\begin{minipage}{0.97\linewidth}\footnotesize
These are deviations from the Moore-Berg human source reference, not errors against an observed Twin-2K human target effect.
\end{minipage}
\end{table}

Prior digital twin studies create a reasonable expectation that richer respondent histories should improve prediction. In one direct comparison, \citet{park2024} report that agents using interviews and surveys reach 86\% of participants' test-retest consistency, compared with 74\% for demographics-only agents. These benchmarks evaluate held-out answers at the individual level, whereas Table~\ref{tab:prompt-calibration} evaluates cross-party population contrasts and their transported calibration. Richer histories may therefore improve average individual-response accuracy without improving the particular group contrast that defines a downstream estimand. In our setting, the free-text persona helps some models but hurts others, so more information is not a reliable design rule for population-level inference. The informational richness of a persona and its inferential usefulness are therefore distinct properties.

Calibration improves many raw differences across prompt regimes, but no persona representation dominates across models. Table~\ref{tab:prompt-calibration} shows that GPT-5.4, GPT-4.1-mini, and the two Qwen models all reach single-digit calibrated deviation in their best reported regimes, while Llama 4 Scout remains much further from the source reference and its party-only calibration moves in the wrong direction. More respondent information therefore does not consistently produce the kind of estimand-specific signal needed for statistical substitutability. Figure~\ref{fig:appendix-moore-berg-components} in Appendix~\ref{sec:oa-components} shows the underlying humanity-rating components for all six models in the common-fields regime.

Across the two empirical settings, the same distinction recurs. Aggregate agreement can improve without paired respondent-level signal, paired signal can exist without correct aggregate replication, and human calibration can reduce bias without producing stable efficiency gains. Improvements in model generation and persona richness also do not move these dimensions in lockstep. The results therefore support our central argument that digital twins should be evaluated by the information they contribute to valid inference about human estimands, not by aggregate behavioral resemblance alone. More broadly, the results reinforce that behavioral realism is neither necessary nor sufficient for statistical substitutability.

\section{Discussion}
This paper asks a practical question. Can LLM digital twin predictions reduce the amount of new human data needed for a behavioral study without weakening the resulting inference? Recent work argues that descriptive similarity is not enough for confirmatory claims. Model predictions must be tied to observed human outcomes through statistical calibration \citep{hullman2026,ludwig2026}. Our results show a twin can reproduce an average human effect and still provide almost no information about which respondents will answer above or below their condition mean.

The results point to three broader conclusions. First, aggregate behavioral fidelity and statistical substitutability are distinct properties. Second, even measurable paired human-twin signal may be too weak or unstable to produce meaningful reductions in human measurement. Third, improvements in model capability, respondent information, and calibration do not reliably increase inferential usefulness. More broadly, these findings suggest that evaluation of AI-generated evidence should move beyond behavioral realism toward inferential usefulness. This distinction is particularly important for IS researchers, whose work frequently relies on surveys, experiments, interviews, digital trace data, and field studies to generate evidence about technology use, digital work, platforms, and human-AI interaction. As AI-generated observations increasingly enter these workflows \citep{gao2025}, methodological attention must shift from whether AI outputs appear realistic to whether they improve inference about the underlying human phenomena being studied.

\subsection{Behavioral fidelity is not statistical substitutability}
The Twin-2K results make the central distinction especially clear. Framing closely reproduces the standardized human effect, yet its residualized human-twin correlation is near zero. By an aggregate replication criterion, the twin appears successful. By the criterion relevant for reducing future human measurement, it contributes essentially no paired information about who departs from the condition mean. Proportion Dominance shows the reverse possibility. It has the strongest respondent-level signal in the Twin-2K benchmark, but its twin effect points in the wrong direction. Neither aggregate fidelity nor paired signal can therefore serve as a reliable proxy for the other.

This distinction helps explain why conventional validation metrics are insufficient for confirmatory use. LLMs can match group distributions while compressing variation or failing to reproduce the same individual response patterns \citep{tjuatja2024,boelaert2025}. Calibration, specialization, and fine-tuning can improve distributional alignment \citep{cao2025,suh2025,kambhatla2026}, but those gains do not necessarily create useful paired prediction. A digital twin pipeline should therefore be evaluated against the human estimand and intended research use rather than through a general claim that the model is or is not human-like.

The finite-sample results sharpen this point. A nonzero residualized correlation can imply a positive theoretical efficiency ceiling without producing an achievable reduction in human measurement. The Twin-2K benchmark shows that estimating the signal from a limited validation sample and working with a finite prediction-only pool can eliminate gains that appear possible under an idealized large-sample calculation. Statistical substitutability is therefore a property of the full design, not simply of a correlation or a model in isolation. More broadly, this finding suggests that behavioral realism and inferential usefulness should be treated as distinct evaluation dimensions for AI-generated evidence.

\subsection{What improves statistical substitutability?}
The Moore-Berg extension was designed as a more favorable test of person-specific prediction because party identification is directly related to the estimands. Respondent-level signal is stronger than in most Twin-2K tasks, which suggests that estimand-relevant respondent information does matter. The gains are still modest, however. This makes it unlikely that the weak Twin-2K results can be explained only by a lack of relevant respondent attributes.

The model comparisons also show that a generally stronger model need not be a better inferential tool for a particular estimand. Later models reduce aggregate source deviation within the GPT, Llama, and Qwen families, but respondent-level signal does not improve monotonically across those comparisons. The same pattern appears for respondent representation. Richer personas sometimes improve agreement and sometimes do not. Prior work likewise shows that prompt design, response format, and fine-tuning can change performance in ways that are not captured by model size or information quantity alone \citep{lutz2025,dominguez2024,ahnert2026,lu2025,binz2025}. The practical lesson is that improvements in general model quality or persona richness should not be assumed to increase statistical substitutability without direct evaluation of the relevant human estimand.

Human calibration provides a second important distinction. Additional labels substantially reduce source-reference deviation for most models, but better aggregate calibration does not consistently produce better inferential efficiency. Human validation can correct error, but it cannot create signal. This explains why large improvements in calibrated agreement can coexist with weak or unstable variance reduction. In some settings, human observations may therefore be more useful for correcting model output than for further fine-tuning it \citep{krsteski2026}.

Collectively, these results suggest that statistical substitutability is difficult to engineer through model quality alone. Improvements in model capability, personalization, and calibration may improve behavioral fidelity without creating the respondent-level signal required for inferential value.

\subsection{Confirmatory substitution versus exploratory design support}

Failure to establish statistical substitutability does not imply that digital twins are useless. Twins with weak paired signal may still support hypothesis generation, treatment screening, instrument development, and other exploratory tasks. This fits the simulate-then-validate workflow described by \citet{hullman2026}. It is also consistent with \citet{golisingh2024}, who find that direct LLM preference elicitation can mislead even when the same tools remain useful for hypothesis generation, and with \citet{gao2025}, who caution against treating LLM outputs as reliable human surrogates without task-specific validation.

Confirmatory mixed-subject use requires stronger evidence. Researchers should define the human estimand and intended use before evaluating a twin pipeline, document the pipeline, generate predictions, and then collect human outcomes for an independent validation sample \citep{abdurahman2025}. The paired human and twin data can be used to estimate average prediction error, measure whether the twins track respondent-level departures from the relevant group mean, and determine the precision that the combined design actually achieves. Reports should state how many human observations were used for pipeline selection, calibration, and final confirmation, along with the number of model predictions, confidence-interval coverage, variance reduction, and effective sample size. Human observations used to tune or choose the simulation procedure are part of the total data cost. A large synthetic sample alone provides little inferential value when the paired signal is weak.

\subsection{Implications for GenAI-enabled research}
Generative AI makes it possible to generate person-specific predictions across new surveys and experimental scenarios, expanding the supply of auxiliary data available to researchers. Yet human observations remain the inferential anchor, and prior work similarly recommends hybrid deployment when domain alignment matters and human labor is costly \citep{wangge2026}. More broadly, these implications extend beyond digital twins. Generative AI is increasingly used to summarize documents, classify text, generate measurements, simulate respondents, and augment empirical analyses. Across these applications, the central methodological question is not whether AI outputs appear plausible, but whether they improve inference about the quantity of interest. Our findings suggest that inferential usefulness should be treated as the primary criterion for evaluating AI-generated evidence.

Statistical substitutability is population specific as well as estimand specific. A calibration relationship that is useful in one sample does not establish usefulness in another. Our Moore-Berg transport analysis illustrates this boundary. The source correction is estimated among Moore-Berg respondents and then applied to predictions for Twin-2K respondents. Without human Moore-Berg outcomes in the Twin-2K target sample, target accuracy and target statistical substitutability cannot be directly validated. Transport therefore requires additional evidence that the human-model error relationship remains sufficiently stable across populations \citep{stuart2011,tipton2013}. The human validation sample must also be trustworthy. Autonomous LLM survey agents can evade common attention checks and can be instructed to distort survey results \citep{westwood2025}. Researchers therefore need procedures that verify respondent identity and protect panel integrity. This threat is different from the transparent use of model predictions as auxiliary data, but it can undermine the human benchmark on which calibration relies.

These issues are particularly important for IS research because many studies rely on surveys, experiments, interviews, digital trace data, and behavioral interventions to generate evidence about technology use, digital work, online behavior, and human–AI interaction. As AI-generated observations become embedded in these workflows, the central methodological challenge shifts from determining whether outputs appear realistic to determining whether they improve inference about the phenomena being studied. Statistical substitutability provides one framework for evaluating when AI-generated observations contribute useful inferential information and function as reliable sources of scientific evidence.

\subsection{Limitations and future research}

Several limitations narrow the scope of the findings. The Twin-2K benchmark evaluates one released baseline pipeline and only matched two-condition estimands. Its retrospective design does not show that the twins would generalize to genuinely new constructs or future interventions. We also cannot determine whether the proprietary models had encountered the classic task materials during training \citep{ludwig2026,hullman2026}. The Moore-Berg extension covers one political domain, six models, and three persona conditions. The efficiency calculations do not include the original cost of collecting the rich respondent profiles. We also hold response-generation method, option order, and extraction fixed. The reported residualized correlations describe the pipelines we evaluated and should not be treated as upper bounds for every possible twin design.

Human outcomes are themselves noisy and context dependent, which places an upper bound on achievable human-twin correspondence. Residualized human-twin correlation therefore reflects both model quality and the predictability of the underlying human outcome. From a design perspective, the requirement is unchanged. A twin must predict enough of the observed human variation to justify the human data and modeling costs associated with using it.

Future work should test a small preregistered set of simulation pipelines across a wider range of estimands and populations. Useful comparisons include restricted generation, open-ended response classification, verbalized distributions, randomized answer order, alternative persona representations, and behavior-specific fine-tuning. Each pipeline should be evaluated on aggregate effects, paired respondent-level signal, variance-reduction ceilings, and finite-sample prediction-assisted performance. When calibration is transported, target human outcomes should be collected whenever possible so that stability can be tested directly. Since analytic choices may favor different validation targets \citep{cummins2025}, pipeline selection should rely on external evidence, a separate calibration sample, or a held-out confirmation sample. Any human observations used for that selection should count toward the total data budget.

\subsection{Conclusion}

LLM-based digital twins should not be treated as generally substitutable or non-substitutable for human participants. Their value for confirmatory research depends on the estimand, the respondent-level signal in the predictions, the human-label budget, the available prediction pool, and the stability of the human-model relationship in the target population. Across the settings studied here, aggregate replication, stronger models, richer personas, and successful calibration do not by themselves establish that twin predictions can reduce human measurement. Evaluating statistical substitutability directly provides a clearer basis for determining when AI-generated observations can function as scientific evidence and when they should be treated as exploratory tools. More broadly, the results suggest that behavioral realism alone is an insufficient standard for evaluating AI-enabled research methods. As generative AI becomes increasingly embedded in empirical research workflows, inferential usefulness should become a primary criterion for evaluating AI-generated evidence. Statistical substitutability provides a practical framework for determining when AI-generated observations can function as scientific evidence rather than merely plausible simulations.

\FloatBarrier
\bibliographystyle{plainnat}
\bibliography{references}

\clearpage
\appendix
\counterwithin{equation}{section}
\counterwithin{table}{section}
\counterwithin{figure}{section}

\section{PPI Variance and Sample Requirement Details}
\label{sec:appendix-ppi-details}

This section gives the standard variance details behind Section~3.2. For independent prediction-only and validation samples, the scalar mean estimator in Equation~\ref{eq:ppi_mean} has variance
\begin{equation}
\operatorname{Var}\!\left(\widehat{\theta}_{\lambda}\right)
=
\frac{\lambda^2\sigma_{\widehat{Y}}^2}{N}
+
\frac{\operatorname{Var}\!\left(Y-\lambda\widehat{Y}\right)}{n}.
\label{eq:ppi_variance}
\end{equation}
Here $\sigma_{\widehat{Y}}^2=\operatorname{Var}(\widehat{Y})$ is the variance of the twin predictions in the prediction-only population. The quantity $\operatorname{Var}(Y-\lambda\widehat{Y})$ is the variance of the prediction-corrected human outcome in the validation population. Let $\sigma_Y^2=\operatorname{Var}(Y)$, $r=\sigma_{\widehat{Y}}/\sigma_Y$, $\rho=\operatorname{Corr}(Y,\widehat{Y})$, and $\kappa=n/N$. Relative to the human-only variance $\sigma_Y^2/n$, the proportional variance reduction is
\begin{equation}
\operatorname{VR}(\lambda)
=
1-\frac{\operatorname{Var}(\widehat{\theta}_{\lambda})}{\sigma_Y^2/n}
=
2\lambda\rho r-\lambda^2r^2(1+\kappa).
\label{eq:ppi_vr}
\end{equation}
The unconstrained variance-minimizing coefficient is
\begin{equation}
\lambda^*
=
\frac{\rho}{r(1+\kappa)}.
\label{eq:ppi_lambda_star}
\end{equation}
Substituting Equation~\ref{eq:ppi_lambda_star} into Equation~\ref{eq:ppi_vr} yields the idealized ceiling in Equation~\ref{eq:ppi_vrmax}. When power tuning restricts $\lambda$ to $[0,1]$, the procedure uses the nearest boundary whenever the unconstrained optimum falls outside that interval. Since $\lambda=0$ returns the human-only estimator, the population-level constrained optimum cannot increase variance relative to human-only estimation.

A fixed unit coefficient does not have the same guarantee. Setting $\lambda=1$ gives
\begin{equation}
\operatorname{VR}(1)
=
2\rho r-r^2(1+\kappa),
\label{eq:ppi_vr_unit}
\end{equation}
which can be negative when prediction variability is too large relative to human variability. This does not conflict with the nonnegative ceiling in Equation~\ref{eq:ppi_vrmax}, which is obtained after optimizing the coefficient rather than fixing it at one.

The effective-sample-size multiplier used in the paper is
\begin{equation}
\operatorname{ESS\ multiplier}(\lambda)
=
\frac{\operatorname{Var}(\widehat{\theta}_{\mathrm{human}})}
{\operatorname{Var}(\widehat{\theta}_{\lambda})}
=
\frac{1}{1-\operatorname{VR}(\lambda)}.
\label{eq:ess-multiplier}
\end{equation}
Under the usual inverse-sample-size approximation, this quantity reports how many human-only observations would be needed to match the precision of the prediction-assisted design.

For the finite-sample planning problem, let $n_L$ denote the number of labeled human observations and let $N_0$ denote the reference human-only sample size. Let $\rho_+$ be the nonnegative estimated residualized correlation. Equating the prediction-assisted precision with the reference human-only precision and solving for the prediction-only sample gives
\begin{equation}
N_{\mathrm{req}}
=
\left\lceil
\frac{n_L(N_0-n_L)}{n_L-N_0(1-\rho_+^2)}
\right\rceil.
\label{eq:nreq}
\end{equation}
This requirement is finite only when the denominator is positive, which is equivalent to the threshold in Equation~\ref{eq:threshold}. In the empirical benchmark, the design must also have at least $N_{\mathrm{req}}$ eligible prediction-only observations remaining after the labeled respondents are removed.
\FloatBarrier

\section{Moore-Berg Transported PPI Coefficients and Efficiency}
\label{sec:appendix-lambda}

In this appendix, $S$ denotes the Moore-Berg source sample, where both human outcomes and model predictions are observed, and $T$ denotes the Twin-2K target sample, where only model predictions are observed. For a given Moore-Berg estimand, $\widehat{\tau}_{Y,S}$ is the survey-weighted human source contrast, $\widehat{\tau}_{\widehat{Y},S}$ is the corresponding model-predicted source contrast, and $\widehat{\tau}_{\widehat{Y},T}$ is the model-predicted target contrast. To keep the derivation compact, write
\begin{equation}
H=\widehat{\tau}_{Y,S}, \qquad
P_S=\widehat{\tau}_{\widehat{Y},S}, \qquad
P_T=\widehat{\tau}_{\widehat{Y},T}.
\end{equation}
The transported estimator studied in the main paper can then be written as
\begin{equation}
\widehat{\tau}_T(\lambda)=H+\lambda(P_T-P_S).
\end{equation}
Across repeated draws of the labeled Moore-Berg source sample, $P_T$ is fixed. The part of the estimator variance that depends on $\lambda$ is therefore
\begin{align}
\operatorname{Var}\!\left[\widehat{\tau}_T(\lambda)\right]
&= \operatorname{Var}(H-\lambda P_S) \notag\\
&= \operatorname{Var}(H)
+\lambda^2\operatorname{Var}(P_S) \notag\\
&\quad -2\lambda\operatorname{Cov}(H,P_S).
\end{align}
When $\operatorname{Var}(P_S)>0$, the variance-minimizing coefficient is
\begin{equation}
\lambda^*
=
\frac{\operatorname{Cov}(H,P_S)}
{\operatorname{Var}(P_S)}.
\label{eq:appendix-lambda-star}
\end{equation}

For a two-arm contrast,
\begin{equation}
H=\bar Y_A^w-\bar Y_B^w,
\qquad
P_S=\overline{\widehat Y}_A^w-\overline{\widehat Y}_B^w,
\end{equation}
where the bars denote survey-weighted arm means. Because the two party arms are disjoint, the covariance and prediction-contrast variance are estimated by adding the corresponding arm-level terms
\begin{align}
\operatorname{Cov}(H,P_S)
&=
\operatorname{Cov}(\bar Y_A^w,\overline{\widehat Y}_A^w)
+
\operatorname{Cov}(\bar Y_B^w,\overline{\widehat Y}_B^w), \\
\operatorname{Var}(P_S)
&=
\operatorname{Var}(\overline{\widehat Y}_A^w)
+
\operatorname{Var}(\overline{\widehat Y}_B^w).
\end{align}

\paragraph{Unit coefficient.}
The unit estimator sets $\lambda=1$. It uses the prediction contrast at full scale and estimates only the human rectification. This is simple and does not spend labeled observations estimating an additional coefficient. It can still increase variance when prediction scale is poorly matched to the human outcome.

\paragraph{Adaptive coefficient.}
The adaptive estimator replaces $\lambda^*$ with an estimate from the current labeled source subset. The same human labels therefore have to estimate the source effect and the covariance ratio in Equation~\ref{eq:appendix-lambda-star}. This estimator is feasible at the stated label budget, but the estimated coefficient can be noisy when the labeled subset is small or the prediction contrast has little stable variance.

\paragraph{Oracle fixed coefficient.}
The oracle fixed estimator computes the same covariance ratio from the complete Moore-Berg source sample and holds it fixed across repetitions. It is not a feasible low-label procedure because it uses human outcomes outside the labeled subset. Its role is diagnostic. It asks how much efficiency the prediction signal could provide if the coefficient were already known with much greater precision.

Table~\ref{tab:model-efficiency} gives the $n_L=115$ cross-model snapshot. The oracle column shows modest available signal for GPT-5.4, GPT-4.1-mini, Qwen3.5, and Qwen3.8. The unit and adaptive columns show that this available signal does not automatically translate into feasible variance reduction.

\begin{table}[!htbp]
\centering
\scriptsize
\begin{threeparttable}
\caption{Moore-Berg source quality and transported efficiency at $n_L=115$}
\label{tab:model-efficiency}
\begin{tabular}{lrrrrr}
\toprule
Model & Source dev. & Mean $\rho_S$ & Unit $\lambda$ VR & Adaptive VR & Oracle fixed VR \\
\midrule
GPT-5.4 & 23.3 & 0.222 & 4.3\% & -7.6\% & 6.0\% \\
GPT-4.1-mini & 44.8 & 0.190 & -1.4\% & -1.3\% & 4.1\% \\
Qwen3.5-27B & 18.0 & 0.195 & -0.4\% & -13.6\% & 5.6\% \\
Qwen3.8-27B & 15.8 & 0.187 & 0.5\% & -17.5\% & 6.9\% \\
Llama 4 Scout & 69.4 & 0.111 & -5.2\% & -308.1\% & 2.2\% \\
Llama 3.1 8B & 80.4 & 0.017 & -32.5\% & -18.9\% & 0.4\% \\
\bottomrule
\end{tabular}
\begin{tablenotes}[flushleft]\footnotesize
\item Notes. Source dev. is mean absolute deviation from the four human Moore-Berg contrasts, where lower is better. Variance reduction uses the common-fields target regime. Unit $\lambda$ fixes the prediction weight at one. Adaptive $\lambda$ is re-estimated in each labeled subset. Oracle fixed $\lambda$ is estimated from the complete source sample and is a mechanism diagnostic, not a feasible low-label estimator.
\end{tablenotes}
\end{threeparttable}
\end{table}
\FloatBarrier

\subsection{Why adaptive weighting can fail under transport}
\label{sec:oa-adaptive-transport}

A nonzero residualized correlation can support useful linear augmentation, but it does not guarantee that either the unit or adaptive estimator will reduce variance. With $\lambda=1$, a prediction that is mis-scaled relative to the human outcome can make the rectification noisier than the human-only estimate. In the source-only unit-coefficient diagnostic, GPT-5.4 is the only model with positive average variance reduction at every budget, ranging from 2.5\% to 5.5\%. GPT-4.1-mini stays close to zero on average. Qwen3.5, Qwen3.8, Llama 4 Scout, and Llama 3.1 8B are usually negative.

The adaptive estimator adds another source of uncertainty. Write
\begin{equation}
\widehat{\lambda}_{\mathrm{ad}}
=
\lambda^*+e_{\lambda},
\end{equation}
where $e_{\lambda}$ is coefficient-estimation error. Holding the realized source and target prediction contrasts fixed, the difference from using the stable coefficient is
\begin{equation}
\widehat{\tau}_T(\widehat{\lambda}_{\mathrm{ad}})
-
\widehat{\tau}_T(\lambda^*)
=
e_{\lambda}(P_T-P_S).
\label{eq:lambda-error-transport}
\end{equation}
The coefficient error is therefore multiplied by the difference between the target and source prediction contrasts. In a same-population setting, the population counterpart of this difference is zero. Under source-to-target transport it can be substantial. A coefficient that would be useful if known precisely can therefore become harmful when it is estimated from a small source sample and applied to a different target population.

Table~\ref{tab:moore-budget} reports the full common-fields budget results. GPT-5.4 is the only model with positive unit-coefficient variance reduction at every reported transported budget. The oracle fixed coefficient remains positive for all six models, although the gains are modest. The adaptive coefficient is negative throughout the budget grid for GPT-5.4, Qwen3.5, Qwen3.8, Llama 4 Scout, and Llama 3.1 8B. GPT-4.1-mini stays near zero. The extreme Llama 4 Scout values are consistent with unstable coefficient estimation because its oracle fixed coefficient remains positive while the adaptive estimator is sharply negative.

\begin{table}[!htbp]
\centering
\scriptsize
\caption{Moore-Berg common-fields variance reduction across human-label budgets}
\label{tab:moore-budget}
\begin{tabular}{lrrrrrr}
\toprule
Model & 25 & 50 & 100 & 115 & 250 & 500 \\
\midrule
\multicolumn{7}{l}{\textit{Panel A. Unit-coefficient variance reduction}} \\
GPT-5.4 & 5.0\% & 1.5\% & 4.4\% & 4.3\% & 5.1\% & 5.1\% \\
GPT-4.1-mini & -1.4\% & -2.5\% & -0.2\% & -1.4\% & 1.7\% & -0.1\% \\
Qwen3.5-27B & -1.8\% & -5.3\% & -2.9\% & -0.4\% & 0.8\% & -0.8\% \\
Qwen3.8-27B & -4.1\% & -9.1\% & -5.0\% & 0.5\% & -0.8\% & -1.8\% \\
Llama 4 Scout & -4.9\% & -8.5\% & -9.6\% & -5.2\% & -5.6\% & -3.3\% \\
Llama 3.1 8B & -29.0\% & -35.4\% & -26.1\% & -32.5\% & -32.1\% & -26.8\% \\
\addlinespace
\multicolumn{7}{l}{\textit{Panel B. Adaptive-coefficient variance reduction}} \\
GPT-5.4 & -4.3\% & -2.6\% & -8.7\% & -7.6\% & -14.7\% & -19.1\% \\
GPT-4.1-mini & -8.2\% & -0.1\% & -0.4\% & -1.3\% & -1.1\% & -2.1\% \\
Qwen3.5-27B & -7.0\% & -9.6\% & -12.2\% & -13.6\% & -16.3\% & -15.3\% \\
Qwen3.8-27B & -15.4\% & -10.6\% & -16.9\% & -17.5\% & -20.9\% & -25.2\% \\
Llama 4 Scout & -308.3\% & -329.1\% & -319.1\% & -308.1\% & -296.0\% & -293.0\% \\
Llama 3.1 8B & -22.9\% & -20.8\% & -16.5\% & -18.9\% & -14.3\% & -11.4\% \\
\addlinespace
\multicolumn{7}{l}{\textit{Panel C. Oracle fixed-coefficient variance reduction}} \\
GPT-5.4 & 6.5\% & 4.7\% & 5.8\% & 6.0\% & 6.5\% & 6.4\% \\
GPT-4.1-mini & 4.4\% & 3.2\% & 4.4\% & 4.1\% & 5.4\% & 4.5\% \\
Qwen3.5-27B & 5.5\% & 4.1\% & 4.8\% & 5.6\% & 6.2\% & 5.2\% \\
Qwen3.8-27B & 5.0\% & 2.9\% & 4.0\% & 6.9\% & 6.0\% & 5.4\% \\
Llama 4 Scout & 2.6\% & 1.3\% & 0.4\% & 2.2\% & 1.3\% & 2.1\% \\
Llama 3.1 8B & 0.7\% & 0.8\% & 0.3\% & 0.4\% & 0.2\% & 0.5\% \\
\bottomrule
\end{tabular}
\begin{minipage}{0.98\linewidth}\footnotesize
Entries are average percent variance reduction across the four Moore-Berg estimands in the transported common-fields analysis. Negative values mean that the estimator is more variable than the human-only estimator. The adaptive coefficient is re-estimated in each labeled subset. The oracle coefficient is estimated from the complete source sample and is included as a mechanism diagnostic rather than a feasible low-label procedure.
\end{minipage}
\end{table}
\FloatBarrier

The full-source coefficients confirm that some useful information is present, but feasible coefficient estimation does not recover it reliably in this transported design. More flexible weighting is not automatically safer when the weight must be learned from a small source sample and the prediction contrast shifts across populations.
\FloatBarrier

\section{Reproducibility and Documentation}
\label{sec:oa-reproducibility}

The Twin-2K benchmark uses 500 repetitions per reference design and fraction, with labeled and prediction-only respondents disjoint by respondent identifier. Requested prediction-only sizes are never truncated to available capacity. The Moore-Berg analyses retain successful parsed records, deduplicate by respondent identifier, and use a common respondent pool when comparing models. Within-family model comparisons reuse the same four source estimands and common evaluation definitions. All point estimates described as transported are conditional on the source-to-target error-stability assumption. No target human outcomes are imputed or treated as observed. Because response-generation and presentation choices are fixed, the reported estimates apply to the procedures evaluated here. The analysis does not verify model-training overlap with the classic task materials and should not be read as a prospective test of entirely new questions.

Appendix~\ref{sec:appendix-genai_doc} provides the GenAI use documentation. It identifies the released Twin-2K predictions, the six Moore-Berg simulation models, the prompt regimes, the generation and validation workflow, the role of Gen-AI in this study, human verification procedures, and observed failure modes. The corresponding code, prompt templates, and paper-generation files are preserved in the linked repositories.
\FloatBarrier

\section{GenAI Use Documentation}
\label{sec:appendix-genai_doc}
This appendix incorporates the GenAI use documentation. It identifies the released Twin-2K predictions, the six Moore-Berg simulation models, the prompt regimes, the generation and validation workflow, the role of GenAI in this study, human verification procedures, and observed failure modes. The corresponding code and prompt templates are preserved in this anonymized repository: \url{https://anonymous.4open.science/r/llm_digital_twin_stat_sub}.

\subsection{Purpose and Scope}

This appendix documents how generative AI was used in the research process for the accompanying manuscript. It separates three roles:

\begin{enumerate}
    \item \textbf{Research infrastructure under evaluation.} LLMs generated person-specific survey predictions that form the auxiliary data used in the proposed mixed-subject research design.
    \item \textbf{Research implementation.} LLM APIs and locally served models were used to generate Moore-Berg source predictions and Twin-2K target predictions under predefined persona regimes.
    \item \textbf{Research assistance.} LLMs supported literature discovery, code review, interpretation checks, and manuscript formating/polishing. Human researchers retained responsibility for study design, empirical verification, theoretical framing, and final claims.
\end{enumerate}

The full code, prompts, analysis scripts, and generated paper artifacts are maintained in the following repository:

\begin{itemize}
    \item Anonymized repository: \url{https://anonymous.4open.science/r/llm_digital_twin_stat_sub/README.md}
    
\end{itemize}

\subsection{Models, Versions, and Access Period}

\small
\begin{longtable}{@{}P{2.35cm}P{3.05cm}P{2.05cm}P{6.1cm}@{}}
\toprule
\textbf{Role} & \textbf{Model or system} & \textbf{Access period} & \textbf{Use in the project} \\
\midrule
\endfirsthead
\toprule
\textbf{Role} & \textbf{Model or system} & \textbf{Access period} & \textbf{Use in the project} \\
\midrule
\endhead
Twin-2K baseline & GPT-4.1-mini baseline released with Twin-2K-500 & 2026 & Existing person-specific predictions were used for the 12-study retrospective benchmark. Generation was performed by the Twin-2K authors, not rerun for this paper. \\
\addlinespace
Moore-Berg & GPT-5.4 & Jul.--Aug. 2026 & Source and target survey-response generation under the common prompt schema. \\
\addlinespace
Moore-Berg & GPT-4.1-mini & Jul.--Aug. 2026 & Source and target survey-response generation and within-family comparison with GPT-5.4. \\
\addlinespace
Moore-Berg & Llama 3.1 8B & Jul.--Aug. 2026 & Open-weight source and target generation through an OpenAI-compatible local serving endpoint. \\
\addlinespace
Moore-Berg & Llama 4 Scout & Jul.--Aug. 2026 & Open-weight source and target generation and within-family comparison with Llama 3.1 8B. \\
\addlinespace
Moore-Berg & Qwen3.5-27B & Jul.--Aug. 2026 & Open-weight source and target generation through the common vLLM-compatible workflow. \\
\addlinespace
Moore-Berg & Qwen3.8-27B & Jul.--Aug. 2026 & Open-weight source and target generation and within-family comparison with Qwen3.5-27B. \\

\bottomrule
\end{longtable}
\normalsize

Run-specific model strings, parameters, output timestamps, and retry records are preserved in the scripts, raw CSV metadata, and repository history. The manuscript reports model families using the labels recorded in the analysis outputs. The within-family comparisons are descriptive model-generation comparisons rather than controlled parameter-scaling experiments.

\subsection{Prompting and Generation Workflow}

\subsubsection{Prompt regimes}

The Moore-Berg source prompt describes one political-survey respondent using party, party-identification strength, ideology, age, gender, race or ethnicity, education, household income, region, state, and employment. The target simulations use three predefined information regimes:

\begin{enumerate}
    \item \texttt{party\_only}: political party only;
    \item \texttt{common\_fields}: a common structured set of demographic and political fields;
    \item \texttt{persona\_summary}: a free-text summary of the respondent's broader Twin-2K profile.
\end{enumerate}

All prompt templates are versioned in \texttt{prompts/moore\_berg/}. Each prompt asks the model to simulate one respondent, answer eight warmth and humanity questions on a 0--100 scale, avoid explanation, and return a fixed JSON object. The same substantive questions and output keys are used across models and persona regimes.

\subsubsection{Generation settings}

Open-weight models were called through an OpenAI-compatible local endpoint. The generic generation script records the model identifier, temperature, top-$p$, parse status, elapsed time, raw output, and error information for every respondent. Its default open-weight settings are temperature 1.0, top-$p$ 1.0, a 512-token output limit, and up to four retries. Actual run-specific values are retained in each output file.

The pipeline first attempts to parse the complete response as JSON and then searches for a JSON object if surrounding text is present. Eight required numeric fields are checked. Numeric outputs are converted to floats and bounded to the intended 0--100 response scale. Derived prejudice and dehumanization quantities are calculated only when all required fields pass validation.

\subsubsection{Retries, completion, and deduplication}

API or serving errors trigger exponential backoff with random jitter. Every attempted response retains an error field and raw output field. Successful respondent identifiers are used to skip completed cases when a job is resumed. Downstream analysis retains successfully parsed records and deduplicates by respondent identifier, keeping the latest valid record when retry rows exist.

\subsection{Human Oversight and Verification}

Human judgment remained central at the following stages:

\begin{itemize}
    \item selecting the behavioral studies and defining each treatment contrast;
    \item checking Twin-2K question mappings and exclusion decisions;
    \item reconstructing the four Moore-Berg human estimands and confirming that they match the published values to rounding;
    \item designing the residualized-correlation, sample-planning, PPI, calibration, and transport analyses;
    \item reviewing scripts, parse failures, duplicate records, and output-completeness checks;
    \item distinguishing source-reference alignment from unobserved target accuracy;
    \item checking numerical claims against saved analysis tables and rerunnable code;
    \item determining the final theoretical framing, limitations, and contribution claims.
\end{itemize}

The simulation models did not choose the estimands, select the final model based on its reported performance, or determine the inferential conclusions. Human outcomes were never replaced by generated outcomes in the benchmark. The model outputs were treated as predictions whose value was evaluated using observed human labels.

For prospective use, the integrity of the labeled validation sample is itself part of the design. Human verification procedures should therefore address the possibility that online panels contain autonomous or assisted synthetic respondents, since contamination of the human anchor would compromise both calibration and evaluation.

\subsection{Failure Modes, Risks, and Mitigations}

\small
\begin{longtable}{P{3.1cm}P{5.05cm}P{5.25cm}}
\toprule
\textbf{Failure mode or risk} & \textbf{Observed or anticipated consequence} & \textbf{Mitigation} \\
\midrule
\endfirsthead
\toprule
\textbf{Failure mode or risk} & \textbf{Observed or anticipated consequence} & \textbf{Mitigation} \\
\midrule
\endhead
Invalid or incomplete structured output & Missing survey fields or unusable responses & Required-key validation, parse flags, retries, raw-output retention, and exclusion of unsuccessful records \\
Out-of-range numeric values & Responses outside the intended 0--100 scale & Numeric validation and bounding to the stated response scale \\
Duplicate retry rows & Multiple records for the same respondent & Deduplication by respondent identifier using the latest valid record \\

Model-version instability & Newer models may improve one validation target while worsening another & Report aggregate, individual, and inferential metrics separately; avoid interpreting model-generation contrasts as parameter-scaling effects \\

Privacy and representational risk & Rich person-specific profiles may reveal sensitive attributes or misrepresent individuals and groups & Use anonymized identifiers, restrict analysis to approved data resources, avoid treating model output as authentic participant speech, and retain human accountability \\
\bottomrule
\end{longtable}
\normalsize

\subsection{Reproducibility and Provenance}

The analysis repository contains the study loaders, prompts, generation scripts, PPI implementation, simulation seeds, paper tables, and model-comparison outputs. The Twin-2K benchmark uses 500 repetitions for each design and label fraction, with labeled and prediction-only samples disjoint by respondent identifier. The Moore-Berg budget analyses use 500 repetitions per estimand and budget. The scripts retain explicit random seeds and save both repetition-level and summary-level outputs.

\subsection{Statement of Responsibility}

GenAI materially enabled the research capability studied in the paper and assisted several stages of research execution. It did not replace human responsibility for evidence selection, statistical design, verification, interpretation, or authorship. The authors retain responsibility for all empirical claims, citations, code, and conclusions.
\FloatBarrier

\section{Twin-2K Study Scope}
\label{sec:appendix-scope}

\begin{table}[!htbp]
\centering
\small
\caption{Primary Twin-2K contrasts}
\label{tab:appendix-scope}
\begin{tabularx}{\textwidth}{lY}
\toprule
Study & Primary contrast \\
\midrule
Outcome Bias & Successful versus failed medical outcome \\
Sunk Cost & Sunk-cost present versus absent \\
Less-is-More & Gamble B versus Gamble A \\
Framing & Loss versus gain frame \\
Absolute / Relative Savings & Calculator versus jacket discount scenario \\
Myside Bias & Ideologically mirrored policy scenario \\
Anchoring & High versus low redwood-height anchor \\
Proportion Dominance & 98\% versus 500-lives frame \\
WTA/WTP & Willingness-to-accept versus willingness-to-pay \\
Allais & Form 1 versus Form 2 risky-choice rate \\
Linda / Conjunction & Bank teller only versus bank teller plus feminist \\
Base Rate & 30-engineer versus 70-engineer condition \\
\bottomrule
\end{tabularx}
\begin{minipage}{0.98\linewidth}\footnotesize
A 95\% versus 500-lives Proportion Dominance contrast is retained as a robustness analysis because it shares the baseline condition with the primary 98\% contrast. Omission Bias and Denominator Neglect are one-sample proportions. Probability Matching is a repeated strategy score. False Consensus and Nonseparability of Risk and Benefit use association estimands. These tasks are excluded because they do not share the common two-condition estimand, not because of model performance.
\end{minipage}
\end{table}
\FloatBarrier
\FloatBarrier

\section{Moore-Berg Estimands and Prompt Regimes}
\label{sec:appendix-moore-berg}

The first two estimands measure how much Democrats overestimate Republican prejudice and dehumanization toward Democrats. The other two measure how much Republicans overestimate Democratic prejudice and dehumanization toward Republicans. Positive values indicate that the perceived out-party bias is larger than the bias reported by the out-party itself. The corresponding human values are 23.95, 33.20, 25.43, and 37.62. The target prompt regimes use party identity alone, a structured common-fields profile, or a free-text persona summary.

\begin{table}[!htbp]
\centering
\scriptsize
\caption{Survey-weighted residualized source correlations by estimand}
\label{tab:appendix-rho}
\begin{tabular}{lrrrr}
\toprule
Model & Dem MP & Dem MD & Rep MP & Rep MD \\
\midrule
GPT-5.4 & 0.256 & 0.187 & 0.235 & 0.212 \\
GPT-4.1-mini & 0.256 & 0.171 & 0.188 & 0.143 \\
Qwen3.5-27B & 0.226 & 0.177 & 0.222 & 0.157 \\
Qwen3.8-27B & 0.224 & 0.186 & 0.168 & 0.172 \\
Llama 4 Scout & 0.094 & 0.082 & 0.164 & 0.104 \\
Llama 3.1 8B & -0.018 & 0.061 & -0.020 & 0.044 \\
\bottomrule
\end{tabular}
\begin{minipage}{0.97\linewidth}\footnotesize
MP denotes meta-prejudice minus the opposing party's actual prejudice. MD denotes the analogous meta-dehumanization contrast. Residualization is performed within the relevant party and outcome component using Moore-Berg survey weights.
\end{minipage}
\end{table}
\FloatBarrier
\FloatBarrier

\section{Finite-Sample Sensitivity}
\label{sec:oa-sensitivity}

\begin{table}[!htbp]
\centering
\scriptsize
\caption{Cohen $d=0.5$ sensitivity at the 90\% human-label fraction}
\label{tab:appendix-sensitivity}
\begin{tabular}{lrrrr}
\toprule
Study & Full $\rho$ & Success rate & Successful reps. & Conditional VR \\
\midrule
Proportion Dominance & 0.260 & 0.178 & 89 & 2.9\% \\
Myside Bias & 0.170 & 0.030 & 15 & 24.1\% \\
Outcome Bias & 0.025 & 0.010 & 5 & -0.9\% \\
WTA/WTP & 0.072 & 0.004 & 2 & -30.5\% \\
Less-is-More & 0.078 & 0.002 & 1 & \NA \\
Base Rate & 0.033 & 0.000 & 0 & \NA \\
Linda / Conjunction & 0.031 & 0.000 & 0 & \NA \\
Absolute / Relative Savings & 0.009 & 0.000 & 0 & \NA \\
Framing & 0.003 & 0.000 & 0 & \NA \\
Allais & 0.000 & 0.000 & 0 & \NA \\
Sunk Cost & -0.000 & 0.000 & 0 & \NA \\
Anchoring & -0.008 & 0.000 & 0 & \NA \\
\bottomrule
\end{tabular}
\begin{minipage}{0.97\linewidth}\footnotesize
All rows use $N_0=128$, $n_L=114$, 500 repetitions, and a minimum planning correlation of approximately 0.331. Success requires a finite required prediction-only size, sufficient remaining disjoint capacity, and successful PPI estimation. Conditional variance reduction is calculated only over successful repetitions and is unstable when success is rare.
\end{minipage}
\end{table}
\FloatBarrier
\FloatBarrier

\clearpage
\newgeometry{margin=0.55in}
\begin{landscape}
\section{Full Moore-Berg Component Diagnostics}
\label{sec:oa-components}

Figure~\ref{fig:appendix-moore-berg-components} reports the underlying humanity-rating components for all six models in the common-fields regime. Human bars provide the source reference, while raw and calibrated bars describe target predictions rather than observed target outcomes.

\begin{center}
\centering
\includegraphics[width=0.98\linewidth,height=0.75\textwidth,keepaspectratio]{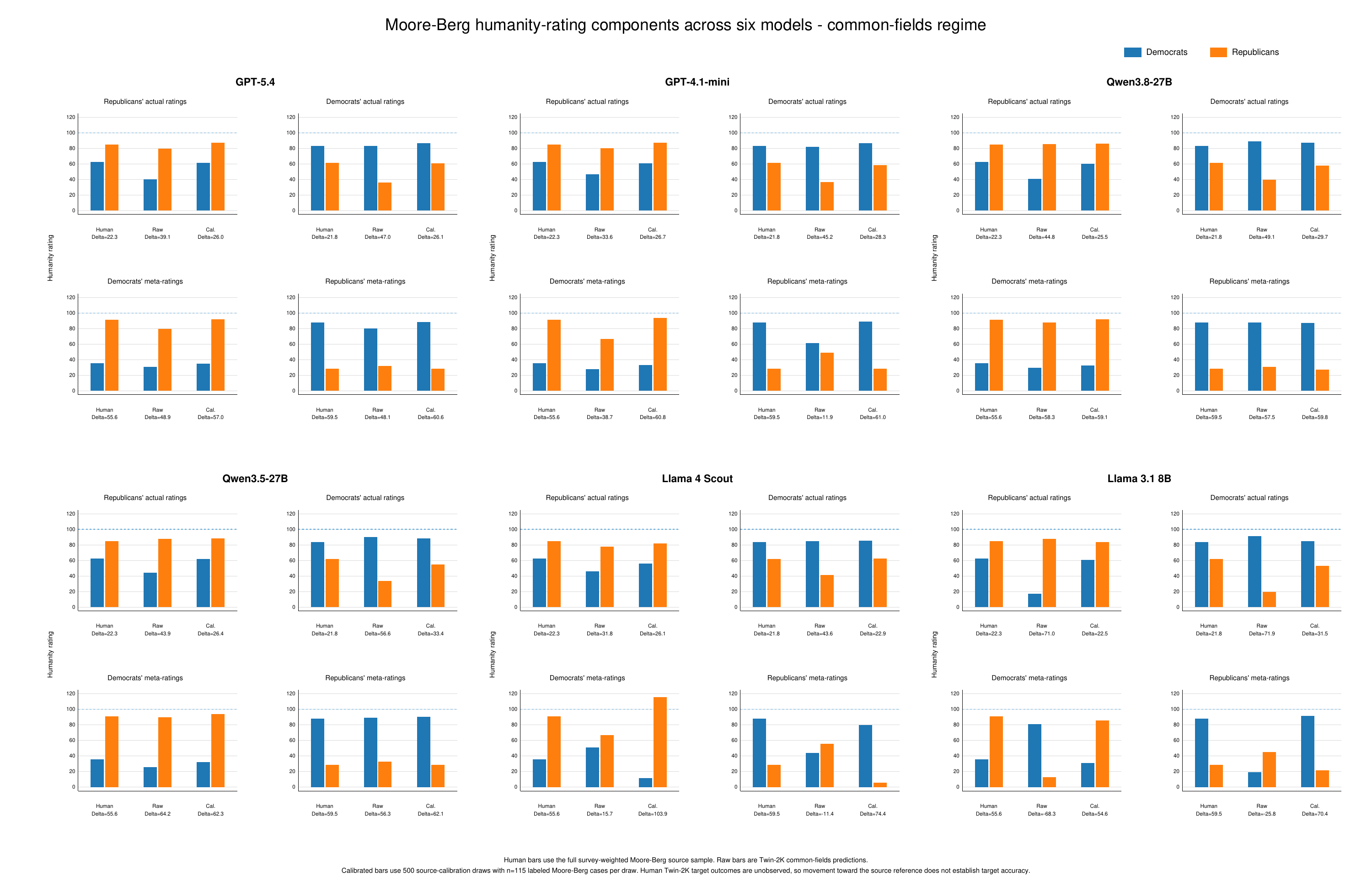}
\captionsetup{hypcap=false}
\captionof{figure}{Humanity-rating components for all six models in the common-fields regime. Human bars report survey-weighted Moore-Berg source means. Raw bars report Twin-2K common-fields predictions. Calibrated bars apply source corrections estimated from 500 draws with 115 labeled Moore-Berg respondents per draw. Each $\Delta$ reports the within-panel party gap. Human outcomes for the Twin-2K target sample are unobserved, so movement toward the source reference does not establish target accuracy.}
\label{fig:appendix-moore-berg-components}
\end{center}
\end{landscape}
\restoregeometry

\end{document}